\documentclass[manuscript,screen]{acmart}

\AtBeginDocument{%
  \providecommand\BibTeX{{%
    \normalfont B\kern-0.5em{\scshape i\kern-0.25em b}\kern-0.8em\TeX}}}

\usepackage{float}
\usepackage{subfig}
\usepackage{graphicx,wrapfig}
\usepackage{amsfonts}
\usepackage{amsmath}
\usepackage{xcolor}
\usepackage[para]{threeparttable}

\graphicspath{{images/}} 
\usepackage{multirow}

\AtBeginDocument{%
  \providecommand\BibTeX{{%
    \normalfont B\kern-0.5em{\scshape i\kern-0.25em b}\kern-0.8em\TeX}}}

\newcommand{\rebut}[1]{\textcolor{black}{#1}}

\begin{document}

\title{An information-theoretic perspective on intrinsic motivation in reinforcement learning: a survey}

\author{Arthur Aubret}
\email{arthur.aubret@liris.cnrs.fr}
\affiliation{
    \institution{Univ Lyon, UCBL, CNRS, INSA Lyon, LIRIS, UMR5205, F-69622 Villeurbanne}
    \city{Lyon}
    \country{France}
}

\author{Laetitia Matignon}
\email{laetitia.matignon@liris.cnrs.fr}
\affiliation{
    \institution{Univ Lyon, UCBL, CNRS, INSA Lyon, LIRIS, UMR5205, F-69622 Villeurbanne}
    \city{Lyon}
    \country{France}
}

\author{Salima Hassas}
\email{salima.hassas@liris.cnrs.fr}
\affiliation{
    \institution{Univ Lyon, UCBL, CNRS, INSA Lyon, LIRIS, UMR5205, F-69622 Villeurbanne}
    \city{Lyon}
    \country{France}
}

\renewcommand{\citet}[1]{\cite{#1}}
\newcommand{\argmax}[1]{\underset{#1}{\operatorname{arg}\,\operatorname{max}}\;}
\newcommand{\argmin}[1]{\underset{#1}{\operatorname{arg}\,\operatorname{min}}\;}
\newcommand{\E}{\mathbb{E}}
\newcommand{\B}{\mathcal{B}}
\renewcommand{\eqref}[1]{Equation \ref{#1}}
\newcommand{\tabref}[1]{Table \ref{#1}}
\newcommand{\secref}[1]{Section \ref{#1}}
\newcommand{\figref}[1]{Figure \ref{#1}}
\newcommand{\appref}[1]{Appendix \ref{#1}}
\newcommand{\chapref}[1]{Chapter \ref{#1}}
\newcommand{\metname}{DisTop}
\newcommand{\Tau}{\mathcal{T}}
\newcommand{\Rho}{\mathrm{P}}
\begin{abstract}
The reinforcement learning (RL) research area is very active, with an important number of new contributions; especially considering the emergent field of deep RL (DRL). However a number of scientific and technical challenges still need to be resolved, amongst which we can mention the \textit{ability to abstract actions} or \textit{the difficulty to explore the environment in sparse-reward settings} which can be addressed by intrinsic motivation (IM). We propose to survey these research works through a new taxonomy based on information theory: we computationally revisit the notions of surprise, novelty and skill learning. This allows us to identify advantages and disadvantages of methods and exhibit current outlooks of research. Our analysis suggests that novelty and surprise can assist the building of a hierarchy of transferable skills that further abstracts the environment and makes the exploration process more robust.
\end{abstract}



\maketitle

\section{Introduction}

In reinforcement learning (RL), an agent learns by trial-and-error to maximize the expected rewards gathered as a result of its actions performed in the environment \cite{sutton1998reinforcement}. Traditionally, an agent maximizes a reward defined according to the task to perform: it may be a score when the agent learns to solve a game or a distance function when the agent learns to reach a goal. The reward is then considered as extrinsic (or as a feedback) because the reward function is provided expertly and specifically for the task. With an extrinsic reward, many spectacular results have been obtained on Atari game \cite{bellemare15} with the Deep Q-network (DQN) \cite{mnih2015human} through the integration of deep learning to RL, leading to deep reinforcement learning (DRL). 

However, despite the recent improvements of DRL approaches, they turn out to be most of the time unsuccessful when the rewards are scattered in the environment, as the agent is then unable to learn the desired behavior for the targeted task \citep{franccois2018introduction}. Moreover, the behaviors learned by the agent are hardly reusable, both within the same task and across many different tasks \citep{franccois2018introduction}. It is difficult for an agent to generalize the learnt skills to make high-level decisions in the environment. For example, such skill could be \textit{go to the door} using primitive actions consisting in moving in the four cardinal directions; or even to \textit{move forward} controlling different joints of a humanoid robot like in the robotic simulator MuJoCo \citep{todorov2012mujoco}. 

On another side, unlike RL, developmental learning \cite{piaget1952origins,cangelosi2018babies,oudeyer2016evolution} is based on the trend that babies, or more broadly organisms, acquire new skill while spontaneously exploring their environment \cite{gopnik1999scientist,barto2013intrinsic}. This is commonly called an intrinsic motivation (IM), which can be derived from an intrinsic reward. This kind of motivation allows to autonomously gain new knowledge and skills, which then makes the learning process of new tasks easier \cite{baldassarre2013intrinsically}. For several years now, IM is increasingly used in RL, fostered by important results and the emergence of deep learning. This paradigm offers a greater learning flexibility, through the use of a more general reward function, allowing to tackle the issues raised above when only an extrinsic reward is used. Typically, IM improves the agent ability to explore its environment, to incrementally learn skills independently of its main task, to choose an adequate skill to be improved and even to create a representation of its state with meaningful properties. In addition, as a consequence of its definition, IM does not require additional expert supervision, making it easily generalizable across environments.


\paragraph{Scope of our review.}
In this paper, we study and group together methods through a novel taxonomy based on information theoretic objectives. This way, \textbf{we revisit the notions of surprise, novelty and skill learning and show that they can encompass numerous works.} Each class is characterized by a computational objective that fits its eventual psychological definition. This allows us to situate/relate a large body of works and to highlight important directions of research. To sum up, this paper investigates the use of IM in the framework of DRL and considers the following aspects:
\begin{itemize}
    \item The role of IM in addressing the challenges of DRL.
    \item Classifying current heterogeneous works through few information theoretic objectives.
    \item \rebut{Exhibit advantages of each class of methods.}
    \item Important outlooks of IM in RL within and across each category.
\end{itemize}


\paragraph{Related works.} The overall literature on IM is huge \citep{barto2013intrinsic} and we only consider its application to DRL and IMs related to information theory. Therefore, our study of IMs is not meant to be exhaustive. Intrinsic motivation currently attracts a lot of attention and several works made a restricted study of the approaches. \citet{colas2020intrinsically}  and \citet{amin2021survey} respectively focus on the different aspects of skill learning and exploration; \citet{baldassarre2019intrinsic} studies intrinsic motivation through the lens of psychology, biology and robotic ; \citet{pateria2021hierarchical} review hierarchical reinforcement learning as a whole, including extrinsic and intrinsic motivations; \citet{linke2020adapting} experimentally compare different goal selection mechanisms. In contrast with these approaches, we study a large part of objectives all based on intrinsic motivation through the lens of information theory. We assume that our work is in line with the work of \citet{schmidhuber2008driven}, which postulates that organisms are guided by the desire to compress the information they receive. However, by reviewing the more recent advances in the domain, we formalize the idea of compression with the tools from information theory.

\rebut{\paragraph{Structure of the paper.}} This paper is organized as follows. As a first step, we discuss RL, define intrinsic motivation and explain how it fits the RL framework (\secref{sec:defs}). Then, we highlight the main current challenges of RL and identify the need for an additional outcome (\secref{sec:defis}). Thereafter, we briefly explain our classification (\secref{sec:classify}), namely surprise, novelty and skill learning and we detail how current works fit it (respectively \secref{sec:infogain}, \secref{sec:novelty} and \secref{sec:skilllearning}). Finally, we highlight some important outlooks of the domain (\secref{sec:outlooks}). 

\section{Definitions and Background}\label{sec:defs}

In this section, we will review the background of RL field explain the concept of IM and described how to integrate IM in the RL framework through goal-parameterized RL, hierarchical RL and information theory. \rebut{We sum up the notations used in the paper in \tabref{tab:notations} in \appref{app:notations}.} 

\subsection{Markov decision process}\label{sec:mdp}

The goal of a Markov Decision Process (MDP) is to maximize the expectation of cumulative rewards received through a sequence of interactions \citep{puterman2014markov}. It is defined by: $S$ the set of possible states; $A$ the set of possible actions; $T$ the transition function $T : S \times A \times S \rightarrow p(s'|s,a)$; $R$ the reward function $R : S \times A \times S \rightarrow \mathbb{R}$; $d_0 : S \rightarrow \mathbb{R}$ the initial distribution of states. An agent starts in a state $s_0$ given by $d_0$. At each time step $t$, the agent is in a state $s_t$ and performs an action $a_t$; then it waits for the feedback from the environment composed of a state $s_{t+1}$ sampled from the transition function $T$, and a reward $r_t$ given by the reward function $R$. The agent repeats this interaction loop until the end of an episode. In reinforcement learning the goal can be to maximize the expected discounted reward defined by $\sum_{t=0}^{\infty} \gamma^t r_t$ where $\gamma \in[0,1]$ is the discount factor. When the agent does not access the whole state, the MDP can be extended to a Partially Observable Markov Decision Process (POMDP) \citep{kaelbling1998planning}. In comparison with a MDP, it adds a set of possible observations $O$ which defines what the agent can perceive and an observation function $\Omega: S \times O \rightarrow \mathbb{R}$ that defines the probability of observing $o \in O$ when the agent is in the state $s$, \textit{i.e} $\Omega(s,o) = p(o|s)$.

A reinforcement learning algorithm aims to associate actions $a$ to states $s$ through a policy $\pi$. This policy induces a t-steps state distribution that can be recursively defined as:
\begin{equation}
d^{\pi}_t(S) = \int_S d^{\pi}_{t-1}(s_{t-1}) \int_A p(s_t|s_{t-1},a)\pi(a|s_{t-1}) da\, ds_{t-1}\label{eq:dpi}
\end{equation}
with $d^{\pi}_0(S)=d_0$. The goal of the agent is then to find the optimal policy $\pi^*$ maximizing the reward:
\begin{equation}
\pi^* = \argmax{\pi} \E_{\substack{s_0\sim d_0(S)\\
a_t \sim \pi(\cdot|s{_t})\\
s_{t+1}\sim p(\cdot|s_t,a_t)}} 
\left[\sum_{t=0}^{\infty} \gamma{^t} R(s{_t},a_t,s_{t+1})\right]
\end{equation}

\rebut{where} $x \sim p(\cdot)$ \rebut{is equivalent to} $x \sim p(x)$.
In order to find the action maximizing the long-term reward in a state $s$, it is common to maximize the expected discounted gain following a policy $\pi$ from a state, noted $V_{\pi}(s)$, or from a state-action tuple, noted $Q_{\pi}(s,a)$ (cf. \eqref{eq:espeQ}). It enables to measure the impact of the state-action tuple in obtaining the
cumulative reward \cite{sutton1998reinforcement}. 
\begin{equation}
Q_{\pi}(s,a) = \E_{\substack{a{_t}\sim\pi(\cdot|s{_t})\\
s_{t+1}\sim p(\cdot|s_t,a_t)}} 
\left[\sum_{t=0}^{\infty} \gamma{^t} R(s{_t},a{_t},s_{t+1})|_{s_0=s,a_0=a} \right]. \label{eq:espeQ}
\end{equation}

To compute these values, one can take advantage of the Bellman equation verified by the optimal Q-function:
\begin{equation}
\label{eq:bellman}
Q^*(s_t,a_t) = \E_{s_{t+1}\sim p(\cdot|s_t,a_t)} \big[ R(s_t,a_t,s_{t+1}) + \gamma \: \max_a Q^*(s_{t+1},a) \big].
\end{equation}

$Q$ and/or $\pi$ are often approximated with neural networks when the state space is continuous or very large \cite{mnih2016asynchronous,lillicrap2015continuous}.

\subsection{Definition of intrinsic motivation}\label{sec:defint}

Simply stated, intrinsic motivation is about doing something for its inherent satisfaction rather than to get a positive feedback from the environment \cite{ryan2000intrinsic}. Looking at this definition, one can notice that intrinsic motivation is defined by contrast with extrinsic motivation; it highlights the difference between the two paradigms. Intrinsic motivation assumes the agent learns on its own while extrinsic motivation assumes there exits an expert/need that supervises the learning process.

According to \citet{singh2010intrinsically}, evolution provides a general intrinsic motivation (IM) function that maximizes a fitness function based on the survival of an individual. Curiosity, for instance, does not immediately produce selective advantages but enables the acquisition of skills providing by themselves some selective advantages. More widely, the use of intrinsic motivation allows to obtain intelligent behaviors which may later serve goals more efficiently than with only a standard reinforcement \cite{baldassarre2013intrinsically,baldassarre2011intrinsic,lehman2008exploiting}. Typically, a student doing his mathematical homework because he/she thinks it is interesting is intrinsically motivated whereas his/her classmate doing it to get a good grade is extrinsically motivated \cite{ryan2000intrinsic}. In this future, the intrinsically motivated student may be more successful in math than the other one. This questions the relevance of using only standard reinforcement methods.


More rigorously, \citet{oudeyer2008can} explain that an activity \textit{is intrinsically motivating for an autonomous entity if its interest depends primarily on the collation or comparison of information from different stimuli and independently of their semantics}. At the opposite, an extrinsic reward results of an unknown environment static function which does not depend on previous experience of the agent on the considered environment. The main point is that the agent must not have any \textit{a priori} on the semantic of the observations it receives. Here the term \textit{stimuli} does not refer to sensory inputs, but more generally to the output of a system which may be internal or external to the independent entity, thereby including \textit{homeostatic} body variables (temperature, hunger, thirst, attraction to sexual activities \dots) \cite{baldassarre2011intrinsic,berlyne1965structure}. Broadly speaking, the motivation of an agent can be internal (\textit{source of motivation}) while still being extrinsic (\textit{why} of the actions). For instance, when an agent is looking for food because of the hunger, hunger is a stimuli coming to the cognitive system of the agent such that it is an internal but extrinsic motivation. As an other example, a child may do his/her home-works because he/she thinks it will be crucial to latter get a job. While the source of the motivation is internal, the true outcome comes from the environment.

Now that the we clarified the notion of intrinsic motivation, we study how to integrate intrinsic motivation in the RL framework.
An extensive overview of IM can be found in \citet{barto2013intrinsic}.

\subsection{A model of RL with intrinsic rewards}\label{sec:modelRL}

Reinforcement learning is derived from behaviorism \cite{skinner} and usually uses extrinsic rewards \cite{sutton1998reinforcement}. However \citet{singh2010intrinsically} and \citet{barto2004intrinsically} reformulated the RL framework to incorporate IM. We can differentiate \textit{rewards}, which are events in the environment, and \textit{reward signals} which are internal stimulis to the agent. Thus, what is named \textit{reward} in the RL community is in fact a \textit{reward signal}. Inside the \textit{reward signal} category, there is a distinction between \textit{primary reward signals} and \textit{secondary reward signals}. The \textit{secondary reward signal} is a local \textit{reward signal} computed through expected future rewards and is related to the value function 
whereas the \textit{primary reward signal} is the standard \textit{reward signal} received from the MDP.

\begin{wrapfigure}{r}{0.3\linewidth}
\begin{centering}
\includegraphics[width=1\linewidth]{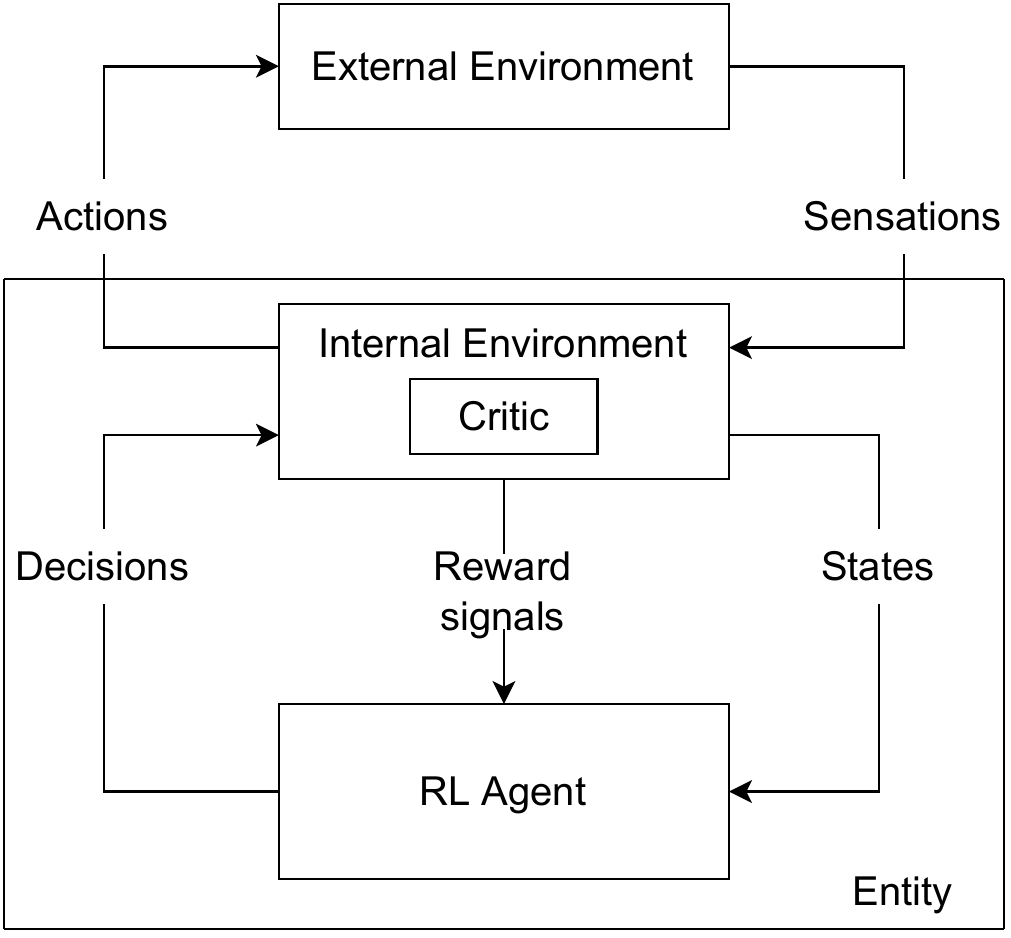}
\caption{\rebut{Model of RL integrating IM}, taken in \protect\citet{singh2010intrinsically}. The environment is factored into an internal and external environment, with all reward coming from the former.}
\label{im:rlintrinsic}
\end{centering}
\end{wrapfigure}

In addition, rather than considering the MDP environment as the environment in which the agent achieves its task, it suggests that the MDP environment can be formed of two parts: the \textbf{external part} which corresponds to the potential task and the environment of the agent; the \textbf{internal part} which computes the MDP states and the \textit{secondary reward signal} using potentially previous interactions. Consequently, we can consider an intrinsic reward as a \textit{reward signal} received from the MDP environment. The MDP state is no more the external state but an internal state of the agent. However, from now, we will follow the terminology of RL and the term \textit{reward} will refer to the \textit{primary reward signal}.


Figure \ref{im:rlintrinsic} summarizes the framework: the critic is in the internal part of the agent, it computes the intrinsic reward and deals with the credit assignment. The agent can merge intrinsic rewards and extrinsic rewards in its internal part. The state includes sensations and any form of internal context; in this section we refer to this state as a contextual state. The decision can be a high-level decision decomposed by the internal environment into low-level actions. 

This conceptual model incorporates intrinsic motivations into the formalism of MDP. Now, we will review how this model is instantiated in practice. Indeed it is possible to extend RL to incorporate the three new components that are intrinsic rewards, high-level decisions and contextual states. We separately study them in the following sections.

\subsection{Intrinsic rewards and information theory}

Throughout our definition of intrinsic motivation, one can notice that the notion of \textit{information} comes up a lot. This is not hazardous and quantifying information proves useful to generate intrinsic rewards. In this section, we provide the basics about information theory and explain how to combine intrinsic and extrinsic rewards. However, we emphasize that intrinsic rewards are not restricted to information measures and their characterization mostly depends on whether the reward function fits the properties of an intrinsic motivation. 

The Shannon entropy quantifies the mean necessary information to determine the value of a random variable. Let $X$ be a random variable with a law of density $p(X)$ satisfying the normalization and positivity requirements, we define its entropy by:
\begin{equation}
H(X) = -\int_{X} p(x)\log p(x) dx .
\end{equation}

In other words, it allows to quantify the disorder of a random variable. The entropy is maximal when $X$ follows a uniform distribution, and minimal when $p(X)$ is equal to zero everywhere except in one value, which is a Dirac distribution. From this, we can also define the entropy conditioned on a random variable $S$. It is similar to the classical entropy and quantifies the mean necessary information to find $X$  knowing the value of an other random variable $S$:
\begin{equation}
H(X|S) = -\int_{S} p(s)\int_{X} p(x|s)\log p(x|s) dx ds.
\end{equation}

The mutual information allows to quantify the information contained in a random variable $X$ about an other random variable $Y$. It can also be viewed as the decrease of disorder brought by a random variable $Y$ on a random variable $X$. The mutual information is defined by:
\begin{equation}
I(X;Y) =  H(X) - H(X|Y)\label{eq:MI}
\end{equation}

We can notice that the mutual information between two independent variables is zero (since $H(X|Y)=H(X)$). Similarly to the conditional entropy, the conditional mutual information allows to quantify the information contained in a random variable about an other random variable, knowing the value of a third one. It can be written in various ways:
\begin{subequations}
\begin{align}
    I(X;Y|S) &= H(X|S) - H(X|Y,S) = H(Y|S) - H(Y|X,S)   \label{information2} \\
    &= D_{KL} \Big[ p(X,Y|S) || p(X|S)p(Y|S)\Big] \label{kldiv} 
\end{align}
\end{subequations}

We can see with \eqref{information2} that the mutual information is symmetric and that it characterizes the decrease in entropy on X brought by Y (or inversely). \eqref{kldiv} defines the conditional mutual information as the Kullback-Leibler divergence \cite{cover2012elements}, \rebut{noted $D_{KL}(.||.)$}, between distribution $P(Y,X|S)$ and the same distribution if $Y$ and $X$ were independent variables (the case where $H(Y|X,S) = H(Y|S)$).

For further information on these notions, the interested reader can refer to \citet{cover2012elements}. Sections 5, 6, 7 illustrate how we can use information theory to reward an agent. In practice, there are multiple ways to integrate an intrinsic reward into a RL framework. The main approach is to compute the agent's reward $r$ as a weighted sum of an intrinsic reward $r_{int}$ and an extrinsic reward $r_{ext}$: $r=\alpha r_{int} + \beta r_{ext}$ \cite{kakade2002dopamine,burda2018exploration}. Of course, one of the weighting coefficient $\alpha$ and $\beta$ can be set to 0. 

\subsection{Decisions and hierarchical RL}\label{sec:hrl}

Hierarchical reinforcement learning (HRL) architectures are adequate candidates to model the decision hierarchy of an agent \cite{barto2003recent,dayan1993feudal,sutton1999between}. \citet{dayan1993feudal} introduced the feudal hierarchy, called \textit{Feudal reinforcement learning}. In this framework, a manager selects the goals that workers will try to achieve by selecting low-level actions. Once the worker achieved the goal, the manager can select an other goal, so that the interactions keep going. The manager rewards the RL-based worker to guide its learning process; we formalize this with intrinsic motivation in the next section. Below, \figureautorefname~\ref{im:abstract_actions} illustrates the use of a hierarchical decision in contrast with the use of low-level actions. At the origin, the hierarchical architectures have been introduced to make easier the long-term credit assignment \cite{dayan1993feudal,sutton1999between}. This problem refers to the fact that rewards can occur with a temporal delay and will only very weakly affect all temporally distant states that have preceded it, although these states may be important to obtain that reward. Indeed, the agent must propagate the reward along the entire sequence of actions (through \eqref{eq:bellman}) to reinforce the first involved state-action tuple. This process can be very slow when the action sequence is large. This problem also concerns determining which action is decisive for getting the reward, among all actions of the sequence. In contrast, if an agent can take advantage of temporally-extended actions, a large sequence of low-level actions become a short sequence of time-extended decisions that make easier the propagation of rewards.


This goal setting mechanism can be extended to create managers of managers so that an agent can recursively define increasingly abstract decisions as the hierarchy of RL algorithms increases. Relatively to \figref{im:rlintrinsic}, the internal environment of a RL module becomes the lower level module. We can model these decisions as \textit{options}. An \textit{option} $op \in \mathcal{O}$ is defined through 3 components: 1- A set of starting states $\mathcal{I} \subset S$ from which an \textit{option} can be applied; 2- A policy (or worker) that is responsible of achieving the \textit{options} with lower-level actions. This is studied in the next section; 3- A completion function $\mathcal{F}$ that specifies the probability of completing the \textit{option} in each state.

Typically, the starting state can derive from $d_0$ (all \textit{options} start at the beginning of an episode) or the full set of states $S$ (\textit{options can start everywhere}). The completion function can also set a probability $0$ everywhere \cite{eysenbach2018diversity}, in this case, it ends at the same time as an episode. Such specific cases often occur \cite{eysenbach2018diversity}. \textit{Options} where originally learnt during a pre-training phase with exclusively extrinsic rewards \cite{sutton1999between}, it was meant to take advantage of expert knowledge on the task. However, in our framework, we are interested in intrinsically motivated agent, so, in the next section, we take a closer look on how to learn the policies that learn to achieve goals using intrinsic motivation. In particular, we will define goals, skills and explain how to build a contextual state.

\subsection{Goal-parameterized RL}\label{sec:goalpam}

Usually, RL agents solve only one task and are not suited to learn multiple tasks. Thus, an agent is unable to generalize across different variants of a task. For instance, if an agent learns to grasp a circular object, it will not be able to grasp a square object. In the developmental model described in \secref{sec:modelRL}, the decisions can be hierarchically organized into several levels where an upper-level takes decision (or sets goals) that a lower-level has to satisfy. This questions: 1- how a DRL algorithm can make its policy dependent on the goal set by its upper-level decision module ? 2- How to compute the intrinsic reward using the goal ? These issues rise up a new formalism based on developmental machine learning \cite{colas2020intrinsically}.

In this formalism, a \textbf{goal} is defined by the pair $(g,R_G)$ where $G \subset \mathbb{R}^d$, $R_G$ is a goal-conditioned reward function and $g \in G$ is the $d\text{-dimensional}$ goal embedding. This contrasts with the notion of task which is proper to an extrinsic reward function assigned by an expert to the agent. With such embedding, one can generalize DRL to multi-goal learning, or even to every available goal in the state space, with the Universal Value Function Approximator (UVFA) \cite{schaul2015universal}. UVFA integrates, by concatenating, the state goal embedding $g$ with the state of the agent to create a contextual state $c = (g,s)$. Depending on the semantic meaning of a skill, we can further enhance the contextual states with other actions or states executed after starting executing the skill (cf. \secref{sec:skilllearning}). 

 We can now define the \textbf{skill} associated to each goal as the goal-conditioned policy $\pi^g(a|s)=\pi(a|g,s)$; in other words, a skill refers to the sensorimotor mapping that achieve a goal \cite{thill2013theories}. This skill may by learnt or unlearnt according to the expected intrinsic rewards it gathers. It implies that, if the goal space is well-constructed (as often a ground state space for example, $R_G=S$), the agent can generalize its policy across the goal space, \textit{i.e} the corresponding skills of two close goals are similar. For example, let us consider an agent moving in a closed maze where every position in the maze can be a goal. We can set $G=S$ and set the intrinsic reward function to be the euclidean distance between the goal and the current state of the agent $R_G: S \times G \rightarrow \mathbb{R}, (s,g) \rightarrow ||s-g||_2$. 

This formalism completes the instantiation of the architectures described in \secref{sec:modelRL}. Now we will explain how, in practice, one can efficiently learn the goal-conditioned policy.

\subsection{Efficient learning with goal relabelling}\label{sec:relabeling}

When the goal space is a continuous state space, it is difficult to determine whether a goal is reached or not, since two continuous values are never exactly equal. Hindsight experience replay (HER) \cite{andrychowicz2017hindsight} tackles this issue by providing a way to learn on multiple objectives with only one interaction. With author's method, the agent can use an interaction done to accomplish one goal to learn on an other goal, by modifying the associated intrinsic reward. This mechanism greatly improves the sample efficiency since it avoids to try all interactions for every goals.

Let us roll out an example. An agent acts in the environment to gather a tuple $(s,s',r_g,a,g)$ where $r_g$ is the reward associated to the goal $g$. The agent can learn on this interaction, but can also use this interaction to learn other goals; to do so, it can change the goal into a new goal and recompute the reward, resulting in a new interaction $(s,s',r_{g'},a,g')$. The only constraint for doing this is that the reward function $R(s,a,s',g')$ has to be known, which is the case with an intrinsic reward function. Typically, an agent can have a goal state and a reward function which is $1$ if it is into that state and $0$ otherwise. At every interaction, it can change its true goal state for its current state and learn with a positive reward.




\section{Challenges of DRL}\label{sec:defis}

In this section, we detail two main challenges of current DRL methods that are partially addressed by IMs.

\subsection{Sparse rewards} \label{sec:sparse}

Classic RL algorithms operate in environments where the rewards are \textbf{dense}, \textit{i.e.} the agent receives a reward after almost every completed action. In this kind of environment, naive exploration policies such as $\epsilon$-greedy \cite{sutton1998reinforcement} or the addition of a Gaussian noise on the action \cite{lillicrap2015continuous} are effective. More elaborated methods can also be used to promote exploration, such as Boltzmann exploration \cite{cesa2017boltzmann,mnih2015human} or an exploration in the parameter-space \cite{plappert2017parameter,ruckstiess2010exploring,fortunato2017noisy}. In environments with \textbf{sparse} rewards, the agent receives a reward signal only after it executed a large sequence of specific actions. The game \textit{Montezuma's revenge} \cite{bellemare15} is a benchmark illustrating a typical sparse reward function. In this game, an agent has to move between different rooms while picking up objects (it can be keys to open doors, torches, ...). The agent receives a reward only when it finds objects or when it reaches the exit of the room. Such environments with sparse rewards are almost impossible to solve with the above mentioned \textit{undirected} exploration policies \cite{thrun1992efficient} since the agent does not have local indications on the way to improve its policy. Thus the agent never finds rewards and cannot learn a good policy with respect to the task \cite{mnih2015human}. Figure \ref{im:sparse_reward} illustrates the issue on a simple environment. 

This issue stresses out the need for \textit{directed} exploration methods \cite{thrun1992efficient}. While intrinsic motivation can provide such direction, the principle of "optimism in face of uncertainty" \cite{audibert2007tuning} can also execute a directed exploration without intrinsic motivation \cite{thrun1992efficient}. Briefly, this principle can incite agents to go in areas with a lot of epistemic uncertainties about its Q-values \cite{ciosek2019better,pacchiano2020optimism}. Yet, it is hard to approximate the epistemic uncertainty and it only slightly improves exploration \cite{ciosek2019better}. This principle can also relate with some intrinsic motivations when we consider uncertainty about models (see \secref{sec:infogainforward}).

\begin{figure}
\begin{centering}
\includegraphics[width=10cm]{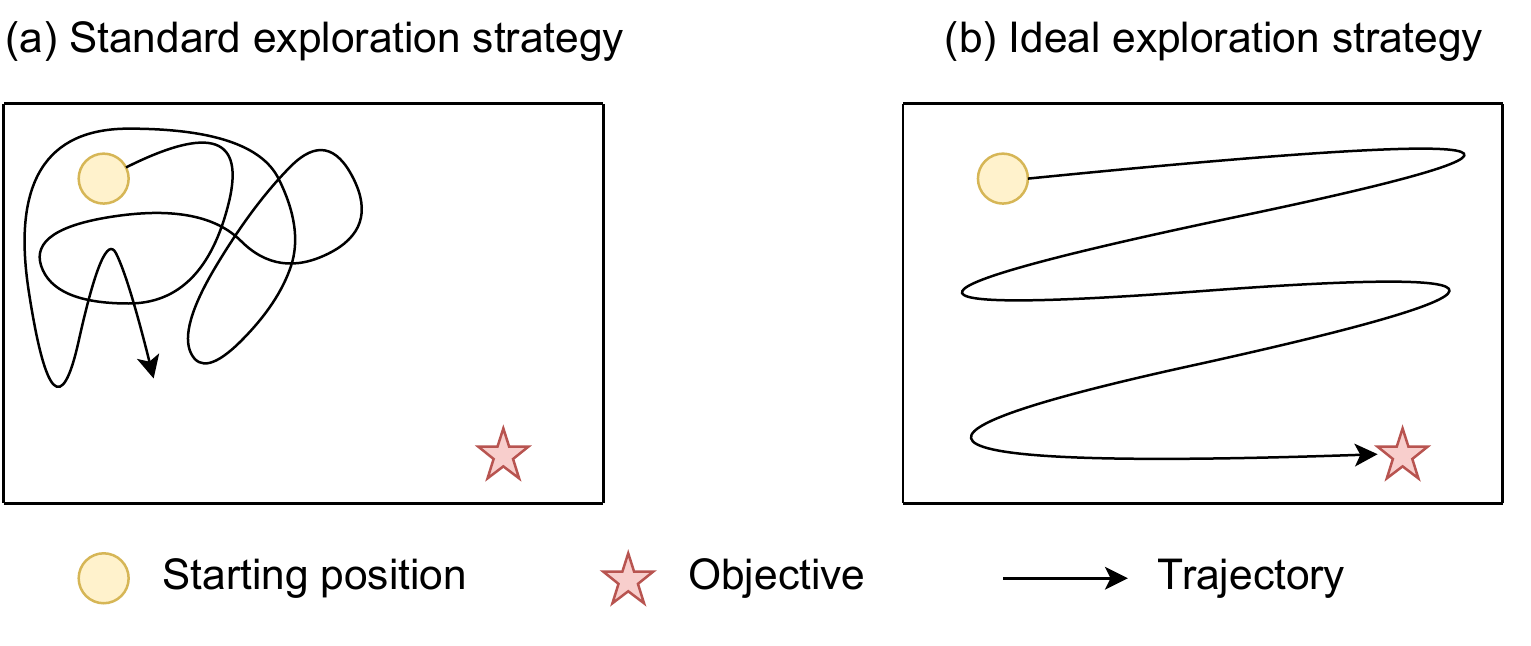}
\caption{\rebut{Example of a very simple sparse reward environment, explored by two different strategies}. The agent, represented by a circle, strives to reach the star. The reward function is one when the agent reaches the star and zero otherwise. (a) the agent explores with standard methods such as $\epsilon\text{-greedy}$; as a result, it stays in its surrounded area because of the temporal inconsistency of its behaviour. (b) we imagine an ideal exploration strategy where the agent covers the whole state space to discover where rewards are located. \rebut{The fundamental difference between the two policies is the volume of the state space explored for a given time.}}
\label{im:sparse_reward2}
\end{centering}
\end{figure}

Rather than working on an exploration policy, it is common to shape an intermediary dense reward function that adds to the reward associated to the task in order to make the learning process easier for the agent \cite{su2015reward}. However, the building of a reward function often reveals several unexpected errors \cite{ng1999policy,amodei2016concrete} and most of the time requires expert knowledge. For example, it may be difficult to shape a local reward for navigation tasks. Indeed, one has to be able to compute the shortest path between the agent and its goal, which is the same as solving the navigation problem. On the other side, the automation of the shaping of the local reward (without calling on an expert) requires too high computational resources \cite{chiang2019learning}. We will see in \secref{sec:infogain}, \ref{sec:novelty} and \ref{sec:skilllearning} how IM is a valuable method to encourage exploration in a sparse rewards setting.

\subsection{Temporal abstraction of actions} \label{sec:abstraction}

As argued in \secref{sec:hrl}, skills, through hierarchical RL, are a key element to speed up the learning process since the number of decisions to take is significantly reduced when skills are used. In particular, they make easier the \textit{credit assignment}. Skills can be manually defined, but it requires some extra expert knowledge \cite{sutton1999between}. To avoid providing hand-made skills, several works proposed to learn them with extrinsic rewards \cite{bacon2017option,subpolicy2020li}. However, if an agent rather learns skills in a \textit{bottom-up} way, \textit{i.e} with intrinsic rewards rather than extrinsic rewards, learnt skills become independent from possible tasks. This way, skills can be reused across several tasks to improve transfer learning \cite{aubret2020elsim,heess2016learning} and an agent can learn skills even though it does not access rewards, improving exploration when rewards are sparse \cite{machado2017laplacian}. Let us illustrate both advantages.

\begin{figure}
\begin{centering}
\includegraphics[width=0.7\linewidth]{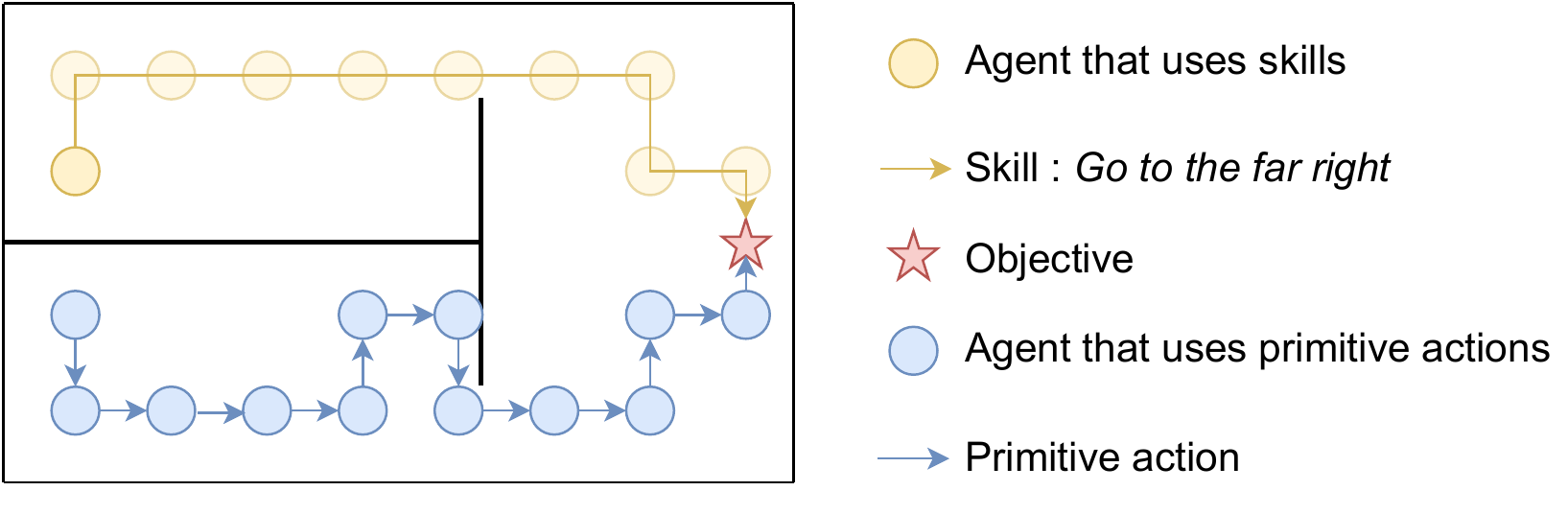}
\caption{\rebut{Example of two policies in a simple environment, one uses \textit{skills} (yellow), the other one only uses primitive actions (blue)}. Agents have to reach the star.}
\label{im:abstract_actions}
\end{centering}
\end{figure}

\paragraph{Exploration when rewards are sparse.} \figref{im:abstract_actions} illustrates the benefit in terms of exploration when an agent hierarchically uses skills. 
The yellow agent can use a skill \textit{Go to the far right}, to reach the rewarding star while the blue agent can only use low-level cardinal movements. 
The problem of exploration becomes trivial for the agent using skills, since one exploratory action can lead to the reward. In contrast, it requires an entire sequence of specific low-level actions for the other agent to find the reward. This problem arises from the minimal number of specific actions needed to get a reward (see also \secref{sec:sparse}). A thorough analysis of this aspect can be found in \cite{nachum2019does}.

\paragraph{Reusing skills across several tasks.} Skills learnt with intrinsic rewards are not specific to a task. Assuming an agent is required to solve several tasks in a similar environment, \textit{i.e} a single MDP with a changing extrinsic reward function, an agent can execute its discovered skills to solve all tasks. Typically, in \figref{im:abstract_actions}, if both agents learnt to reach the star and we move the star somewhere else in the environment, the yellow agent would still be able to execute \textit{Go to the far right} and executing this skill may make the agent closer to the new star. In contrast, the blue agent would have to learn a whole new policy. In \secref{sec:skilllearning}, we provide insights on how an agent can discover skills in a \textit{bottom-up} way.

\section{Classification of methods}\label{sec:classify}

\rebut{In order to tackle the problem of exploration, an agent may want to identify and return in \textbf{rarely visited} states or \textbf{unexpected} states, which can be quantified with current intrinsic motivations. We will particularly focus on two objectives that address the challenge of exploring with sparse rewards, each with different properties: maximizing novelty and surprise. 
Surprise and novelty are specific notions that have often been used in an interchanged way and we are not aware of a currently unanimous definition of novelty \cite{barto2013novelty}. The third notion we study, skill learning, focuses on the issue of skill abstraction. In practice, surprise and novelty are currently maximized as a flat intrinsic motivation, \textit{i.e} without using hierarchical decisions. This mostly helps to improve exploration when rewards are sparse. In contrast, skill learning allows to define time-extended hierarchical skills that enjoy all the benefits argued in \secref{sec:abstraction}.}


Table \ref{tab:taxonomy} sums up our taxonomy.  based on information theory that reflects the high-level studied concepts of novelty, surprise and skill learning. In practice, we mostly take advantage of the \textit{mutual information} to provide a quantity for our conceptual objectives. These objectives are compatible with each other and may be used simultaneously, as argued in \secref{sec:flatim}. Within each category of objectives, we additionally highlight several ways to maximize each objective and provide details about the underlying methods of the literature. We sum up the methods in Tables \ref{tab:surprise}, \ref{tab:novelty} and \ref{tab:skills} and compare their respective advantages when possible.

\begin{table}
\begin{tabular}{|c|c|c|c|}
\hline
 \multicolumn{4}{|c|}{\multirow{2}{*}{\textbf{Surprise}: $I(S';\Phi_T|h,S,A)$, \secref{sec:infogain}} }\\ 
  \multicolumn{4}{|c|}{}
 \\
\hline
Formalism & Information gain  & Information gain  & Information gain  \\ 
 & over forward model & over the true model  & over density model \\
\hline
Sections & \secref{sec:infogainforward} & \secref{sec:predictionerror} & \secref{sec:infogaindensity} \\ 
 \hline
Rewards & $D_{KL}(p(\Phi|h,s,a,s')||p(\Phi|h))$ & $||s' - \hat{s}'||_2^2$  &  $\frac{1}{\sqrt{\hat{N}(s')}}$ \\
\hline
Advantage & \rebut{Simplicity} & \rebut{\textbf{Stochasticity robustness}} & \rebut{Good exploration} \\
\hline 
  \multicolumn{4}{|c|}{\multirow{2}{*}{\textbf{Novelty}: $I(S;Z)$, \secref{sec:novelty}}} 
  \\
  \multicolumn{4}{|c|}{}
  \\
\hline
Formalism & Parametric density  &  \multicolumn{2}{c|}{K-nearest neighbors} \\ 
\hline
Sections & \secref{sec:directdensity} & \multicolumn{2}{c|}{\secref{sec:knearest}}  \\ 
 \hline
Rewards & $- \log \rho(s')$ & \multicolumn{2}{c|}{ $\log (1+ \frac{1}{K} \sum_0^K || f(s') - nn_k(f(S_b),f(s')) ||_2)$ } \\
\hline
\rebut{Advantage} & \rebut{Good exploration} & \multicolumn{2}{c|}{\rebut{\textbf{Best exploration}}}\\
\hline
  \multicolumn{4}{|c|}{\multirow{2}{*}{\textbf{Skill learning}: $I(G; u(\mathcal{T}))$, \secref{sec:skilllearning}}} \\
  \multicolumn{4}{|c|}{}
  \\
\hline
Formalism & Fixed goal distribution  & Goal-state &  Proposing diverse goals \\ 
 & &  achievement & \\
 \hline
 Sections & \secref{sec:predefinedG} & \secref{sec:goalstate} & \secref{eq:diversestate} \\ 
 \hline
Rewards & $\log p(g|s')$ & $-||s_g-s'||_2^2$ & $(1+\alpha_{skew})\log p(s_g)$ \\
\hline
Advantage & \rebut{Simple goal sampling} & \rebut{\textbf{High-granularity skills}} & \rebut{\textbf{More diverse skills}} \\
\hline


\end{tabular}
\caption{Summary of our taxonomy of intrinsic motivations in DRL. The function $u$ outputs a part of the trajectories $\mathcal{T}$, $Z$ and $G$ are internal random variables respectively denoting state representations and self-assigned goals. Please, refer to the corresponding sections for more details about methods and notations. The reward function aims to represent the one used in the category.}
\label{tab:taxonomy}
\end{table}

\section{Surprise}\label{sec:infogain}

In this section, we study methods that maximize the surprise. Firstly, we formalize the notions of surprise, then we will study three approaches for computing intrinsic rewards based of these notions.

\subsection{Definition of surprise}\label{sec:expecsurprise}

In this section, we assume the agent learns either a density model (\secref{sec:infogaindensity}) or a forward model of the environment (Sections \ref{sec:infogainforward} and \ref{sec:predictionerror}) parameterized by $\phi \in \Phi$. The density model induces a marginal distribution of state $p(S|\phi)$ and a forward model computes the next-state distribution conditioned on a tuple state-action $p(S'|S,A,\phi)$. Typically, this can be the parameters of a neural network. Trying to approximate the true model, the agent maintains an approximate distribution $p(\Phi|h)$ of models, where $h_t=h$ refers to the ordered history of interactions $((s_0,a_0,s_1),(s_1,a_1,s_2),\dots, (s_{t-1},a_{t-1},s_t))$. In this section, $h$ simulates a dataset of interactions, we use it to clarify the role of the dataset. It is important to notice that the policy feeds this $h$.

In this case, \textbf{surprise quantifies the mismatch between an expectation and the true experience of an agent} \cite{barto2013novelty,ekman1994nature}. In this paper, we refer to the definition of \citet{itti2009bayesian}, which define it as the discrepancy between a prior distribution of beliefs and the posterior probability distribution following an observation \cite{itti2009bayesian,storck1995reinforcement}. If an agent maximizes the surprise over a model through interactions with the environment, which is often the case \cite{barto2013novelty}, it leads to the expected information gain objective \cite{sun2011planning}. Intuitively, the agent returns in states where it experienced an unexpected transition. Using the KL-divergence to assess the discrepancy, surprise can be computed as $D_{KL}(p(\Phi|h_{t+1})||p(\Phi|h_t))$ where $\phi \in \Phi$ are parameters of a model and $t$ denotes the timestep.

In this case, the agent has a prior distribution about model parameters $p(\Phi)$ and this model can be updated using the Bayes rule:

\begin{equation}
p(\phi|h,s,a,s') = \frac{p(\phi|h)\; p(s'|h,s,a,\phi)}{p(s'|h,s,a)}.
\end{equation}

\paragraph{Information gain over agent's model.} The expected information gain \cite{sun2011planning,little2013learning} over a forward or density model parameterized by $\phi$ can be formulated as:

\begin{subequations}
\begin{align}
	IG(h,A,S',S,\Phi) &= I(S';\Phi|h,A,S) = \E_{\substack{ (s,a) \sim p(\cdot|h) \\ s' \sim p(\cdot | s,a,h)}} D_{KL}(p(\Phi|h,s,a,s')||p(\Phi|h)) \label{eq:trueexpectedinfogain} \\ 
	&\approx \E_{\substack{ (s,a) \sim \pi \\ s' \sim p(\cdot | s,a,h,\phi_T)}} D_{KL}(p(\Phi|h,s,a,s')||p(\Phi|h)) \label{eq:expectedinfogain}
\end{align}
\end{subequations}

 Actively maximizing the expected information gain amounts to reduce the uncertainty of the model. We emphasize that $p(\phi|h) = p(\phi|h,a,s)$ since only full transitions provide information about the true dynamics of the environment. In this case, $p(s'| s,a,h)$ does not refer to the probability induced by the environment, but rather to the probability induced by the current history of transitions. This is stressed out by writing:

\begin{equation}
	p(s'|s,a,h) = \sum_{\phi \in \Phi} p(s'|s,a,h,\phi)p(\phi|s,a,h).\label{eq:marginalphi}
\end{equation}

We highlight that the difference between \eqref{eq:trueexpectedinfogain} and \eqref{eq:expectedinfogain} is important and misleading in the literature \cite{houthooft2016vime,little2013learning,sun2011planning}: in the first equation, the agent imagines new outcomes in order to select actions that maximize the change in the internal model, while in \eqref{eq:expectedinfogain}, the agents acts and uses the new states to update its model.

\paragraph{Information gain over the true forward model.} In our formalism, we assume that there is a distribution of true models $p(\Phi_T)$ that underpins the transition function of the environment $T$. In contrast with $\Phi$, this is a property of the environment. One can see this distribution as a Dirac distribution if only one model exists or as a categorical distribution of several forward models. We define the expected information gain over the true models as:

\begin{subequations}
\begin{align}
    IG(h,A,S',S,\Phi_T) &= I(S';\Phi_T|h,A,S) = H(\Phi_T|h,A,S) - H(\Phi_T|h,A,S,S') \\
        &= \E_{\substack{(s,a) \sim p(\cdot|h),\, \phi_T \sim p(\cdot) \\ s' \sim p(\cdot|s,a,\phi_T)}} \log p(s'|s,a,h,\phi_T) - \log p(s'|s,a,h) \label{eq:predicterror3}.
\end{align}
\end{subequations}
Maximizing \eqref{eq:predicterror3} amounts to look for states that provides new information about the true models distribution. We can see that the left-hand side of \eqref{eq:predicterror3} incites the agent to target inherently deterministic areas, \textit{i.e}, given the true forward model, the agent would exactly know where it ends up. At the opposite, the right-hand term pushes the agent to go in stochastic areas according to its current knowledge. Overall, to improve this objective, an agent has to reach areas that are more deterministic than what it thinks they are. One can see that, assuming $p(s'|s,a,h,\phi_T) \approx  p(s' | s, a, \phi, h)$, one falls back on the expected information gain (see also \eqref{eq:predicterror2}). In contrast with \eqref{eq:expectedinfogain}, this objective takes advantage of the true model, which is most of the time unknown, thereby making the objective hardly tractable. As such, in this perspective, surprise results from an agent-centric approximation of the discrepancy between the agent's model and the environment model.

In the following, we will study three objectives: the expected information gain over the true forward models, the expected information gain over the forward model and the expected information gain over density models.

\subsection{Information gain over the true forward model}\label{sec:predictionerror}


To avoid the need of the true forward model, the agent can omit the left-hand term of \eqref{eq:predicterror3} by assuming the true forward model is modelled as a deterministic forward model. In this case, we can write:

\begin{subequations}
\begin{align}
I(S';\Phi_T|h,A,S) &\propto \E_{\substack{(s,a) \sim p(\cdot|h), \phi_T \sim p(\cdot) \\ s' \sim p(\cdot|s,a,\phi_T)}} - \log p(s'|s,a,h) \label{eq:predicterror4} \\
 &= \E_{\substack{(s,a) \sim p(\cdot|h),\, \phi_T \sim p(\cdot) \\ s' \sim p(\cdot|s,a,\phi_T)}} - \log \sum_{\phi \in \Phi} p(s'|h,s,a,\phi)p(\phi|h) \\
 &\geq \E_{\substack{\phi_T \sim p(\cdot),\, (s,a) \sim p(\cdot|h) \\ s' \sim p(\cdot|s,a,\phi_T), \phi \sim p(\cdot|h)}} - \log p(s'|h,s,a,\phi) \label{eq:predicterror5}
\end{align}
\end{subequations}

where we applied the Jensen inequality in \eqref{eq:predicterror5} and $\phi_T \sim p(\cdot)$ is fixed. One can model $p(s'|h,s,a,\phi)$ with a unit-variance Gaussian distribution in order to obtain a tractable loss. This way, we have:

\begin{subequations}
\begin{align}
	\E_{\substack{(s,a) \sim p(\cdot|h),\, \phi_T \sim p(\cdot) \\ s' \sim p(\cdot|s,a,\phi_T),\, \phi \sim p(\cdot|h)}} - \log p(s' | \phi,h,a,s) &\approx \E_{\substack{(s,a) \sim p(\cdot|h) ,\, s' \sim p(\cdot|s,a,\phi_T) \\ \phi \sim p(\cdot|h),\, \phi_T \sim p(\cdot) }} - \log \frac{1}{(2\pi)^{d/2}}e^{-0.5 (s' - \hat{s}')^T (s' - \hat{s}')} \label{eq:gaussianinfogain} \\
	&\propto \E_{\substack{(s,a) \sim p(\cdot|h) ,\, s' \sim p(\cdot|s,a,\phi_T) \\ \phi \sim p(\cdot|h),\, \phi_T \sim p(\cdot) }} ||s' - \hat{s}'||_2^2 + Const 
\end{align}
\end{subequations}
	%
where 
\begin{equation}
\hat{s}' = \argmax{s'' \in S} p(s''|h,a,s,\phi)
\end{equation}
  represents the mean prediction and $\phi$ parameterizes a deterministic forward model. 
Following the objective, we can extract a generic intrinsic reward as:
\begin{align}
    R(s,a,s')= ||f(s')- f(\hat{s}')||_2^2
    \label{eq:rewpredicterror}
\end{align}

where $f$ is a generic function (e.g. identity or a learnt one) encoding the state space into a feature space. \eqref{eq:rewpredicterror} amounts to reward the predictor error of $\phi$ in the representation $f$. In the following, we will see that learning a relevant function $f$ is the main challenge.

The first natural idea to test is whether a function $f$ is required. \citet{burda2019largescale} learn the forward model from the ground state space and observe it is inefficient when the state space is large. In fact, the euclidean distance is meaningless in such high-dimensional state space. In contrast, they raise up that random features extracted from a random neural network can be very competitive with other state-of-art methods. However they poorly generalize to environment changes. An other model, \textit{Dynamic Auto-Encoder (Dynamic-AE)} \cite{stadie2015incentivizing}, computes the distance between the predicted and the real state in a state space compressed with an auto-encoder \cite{hinton2006reducing}. $g$ is then the encoding part of the auto-encoder. However this approach only slightly improves the results over Boltzmann exploration on some standard Atari games. Other works also consider a dynamic-aware representation \cite{ermolov2020latent}. These methods are unable to handle the local stochasticity of the environment \cite{burda2019largescale}. For example, it turns out that adding random noise in a 3D environment attracts the agent; it passively watches the noise since it is unable to predict the next observation. \label{tele} This problem is also called \textit{the white-noise} problem \cite{pathak2017curiosity,schmidhuber2010formal}. This problem emerges by considering only the right-hand term of \eqref{eq:predicterror3}, making the agent assumes environments are deterministic. Therefore, exploration with prediction error breaks down when this assumption is no longer true.


To tackle exploration with local stochasticity, the \textit{intrinsic curiosity module (ICM)} \cite{pathak2017curiosity} learns a state representation function $f$ end-to-end with an \textit{inverse model} (i.e. a model which predicts the action done between two states). Thus, the function $f$ is constrained to represent things that can be controlled by the agent during next transitions. Secondly, the forward model used in ICM predicts, in the feature space computed by $f$, the next state given the action and the current state. The prediction error does not incorporate the white-noise that does not depend on actions, so it will not be represented in the feature state space. ICM notably allows the agent to explore its environment in the games \textit{VizDoom} and \textit{Super Mario Bros}. Building a similar action space, \textit{Exploration with Mutual Information (EMI)} \cite{pmlr-v97-kim19a} significantly outperforms previous works on Atari but at the cost of several complex layers. EMI transfers the complexity of learning a forward model into the learning of states and actions representation through the maximization of $I([S,A];S')$ and $I([S,S'];A)$. Then, the forward model $\phi$ is constrained to be a simple linear model in the representation space. Furthermore, EMI introduces a \textit{model error} which offloads the linear model when a transition remains strongly non-linear (such as a screen change). However one major drawback of ICM and EMI is the incapacity of their agent to keep in their representation what depends on their long-term control. For instance, in a partially observable environment, an agent may perceive the consequences of its actions several steps later. In addition they remain sensitive to stochasticity when it is produced by an action \cite{burda2019largescale}.

An other way to tackle local stochasticity can be to maximize the improvement of prediction error, or learning progress, of a transition model \cite{schmidhuber1991curious,azar2019world,lopes2012exploration,oudeyer2007intrinsic,kim2020active}. One can see this as approximating the left-hand side of \eqref{eq:predicterror3} with:
\begin{align}
    \log p(s'|s,a,h,\phi_T) - \log p(s'|s,a,h) &\approx \log p(s'|s,a,h') - \log p(s'|s,a,h)
\end{align}

where $h'$ concatenates $h$ with an arbitrary number of additional interactions. As $h'$ becomes large enough and the agent updates its forward model, its forward model converges to the true transition model. Formally, if one stochastic forward model can describe the transitions, we can write:

\begin{subequations}
\begin{align}
    \lim_{|h'|\rightarrow \inf} p(s'|s,a,h') &= \lim_{|h'|\rightarrow \inf} \sum_{\Phi} p(s'|s,a,h',\phi) p(\phi|h') \nonumber \\
    &= p(s'|s,a,h',\phi_T) \label{eq:approxlearningprogress}
\end{align}
\end{subequations}

In practice, we can not wait for discovering a long sequence of new interactions and the reward can be dependent on a small set of interactions and the efficiency of the gradient update of the forward model. Yet, the theoretical connection with the true expected information gain may indeed explain the robustness of learning progress to stochasticity \cite{linke2020adapting}.

\paragraph{Conclusion.} While these methods perform well in deterministic environments, they struggle to offset the determinism assumption that underpines the focus on \eqref{eq:predicterror4}; it results that standard methods focus on the more stochastic areas. Methods that tackle stochasticity may not predict important long-term information about the environment or they need to compute a learning progress measure, which is non-trivial. 

\subsection{Information gain over forward model}\label{sec:infogainforward}
 
In this subsection, we study the works that maximize the expected information gain over forward models. Here, $\phi$ are parameters of a learnt forward model. Using \eqref{eq:expectedinfogain}, we can extract an intrinsic reward:

\begin{equation}
R(s,a,s') = D_{KL}(p(\Phi|h,s,a,s')||p(\Phi|h)).\label{eq:rewinfogain}
\end{equation}

This way, an agent executes actions that provide information about the dynamics of the environment.  This allows, on one side, to push the agent towards areas it does not know, and on the other side to prevent attraction towards stochastic areas. Indeed, if the area is deterministic, environment transitions are predictable and the uncertainty about its dynamics can decrease. At the opposite, if transitions are stochastic, the agent turns out to be unable to predict transitions and does not reduce uncertainty. The exploration strategy \textit{VIME} \cite{houthooft2016vime} computes this intrinsic reward by modelling $p(\phi|h)$ with Bayesian neural networks \cite{graves2011practical}. The interest of Bayesian approaches is to be able to measure the uncertainty of the learned model \cite{blundell2015weight}. This way, assuming a fully factorized Gaussian distribution over model parameters, the KL-divergence has a simple analytic form \cite{houthooft2016vime,linke2020adapting}, making it easy to compute. 
 However, the interest of the proposed algorithm is shown only on simple environments and the reward can be computationally expensive to compute. \citet{achiam2017surprise} propose a similar method (\textit{AKL}), with comparable results, using deterministic neural networks, which are simpler and quicker to apply. The weak performance of both models is probably due to the difficulty to retrieve the uncertainty reduction by rigorously following the mathematical formalism of information gain.


The expected information gain can also be written:
\begin{subequations}
\begin{align}
I(S';\Phi|h,A,S) &= H(S'|h,A,S) - H(S'|A,\Phi,S,h) \nonumber \\
	&\approx - \E_{\substack{(s,a) \sim p(\cdot|h),\, \phi_T \sim p(\cdot)  \\ s' \sim p(\cdot | s,a,h,\phi_T)}} \log p(s'|h,s,a) + \E_{\substack{\phi \sim p(\cdot|h,s,a,s')  \\  (s,a) \sim p(\cdot|h), s' \sim p(\cdot | s,a,h,\phi_T)}} \log p(s' | s, a, \phi, h) \label{eq:predicterror} \\
	&= \E_{\substack{\phi \sim p(\cdot|h,s,a,s'),\, \phi_T \sim p(\cdot) \\  (s,a) \sim p(\cdot|h), s' \sim p(\cdot | s,a,h,\phi_T)}}  - \log \sum_{\phi \in \Phi} p(s'|\phi,h,s,a)p(\phi|h) + \log p(s' | s, a, \phi, h) \label{eq:predicterror2} 
\end{align}
\end{subequations}

 Using similar equations than in \eqref{eq:predicterror2}, in \textit{JDRX} \cite{shyam2018model}, authors show that one can maximize the information gain by computing the Jensen-Shannon or Jensen-Rényi divergence between distributions of states induced by several forward models. The more the models are trained on a state-action tuple, the more they will converge to the expected distribution of next states. Intuitively, the reward represents how much the different transition models disagree on the next-state distribution. Other works also maximize a similar form of disagreement \cite{pathak2019self,yao2021sample,sekar2020planning} by looking at the variance of predictions among several learnt transition models. \rebut{While these models handle the white-noise problem, the main intrinsic issue is computational since they require multiple forward models to train.}

\paragraph{Conclusion.} Despite the theoretical power of the information gain for improving exploration, it remains hard to efficiently estimate it and use it in difficult tasks.

\subsection{Information gain over density model}\label{sec:infogaindensity}

Surprise can also arise by quantifying \textit{the discrepancy between its probability of occurring and the fact that it actually occurred} \cite{barto2013novelty}. To quantify this probability of occuring, in this paragraph, we assume the agent tries to learn a density model $\phi \in \Phi$  that approximates the current marginal density distribution of states $p(s')$. In this setting, we can define the expected information gain over a density model $\rho$ \cite{bellemare2016unifying}:

\begin{align}
IG(h,S,A,S',\Rho)&\approx \E_{\substack{ (s,a) \sim p(\cdot|h),\, \rho_T \sim p(\cdot)  \\ s' \sim p(\cdot | s,a,h,\Rho_T)}} D_{KL}(p(\rho|h,s')||p(\rho|h)).
\end{align}

We hypothesize that the adversarial training that results from the objective (active maximization of the KL-divergence and density fitting) results in an approximately uniform distribution of states (and a uniform density estimation). This may be due to the convexity of the KL-divergence in $p(\rho|h,s')$ and $p(\rho|h)$ but we leave the proof to future work. To our knowledge, no works directly optimize this objective, but it has been shown that the information gain lower-bounds the squared inverse pseudo-count objective \cite{bellemare2016unifying}, which derives from count-based objectives; in the following, we will review \textit{count} and \textit{pseudo-count} objectives.

To efficiently explore its environment, an agent can count the number of times it visits a state and returns in rarely visited states. Such methods are said to be \textit{count-based} \cite{strehl2008analysis}. As the agent visits a state, the intrinsic reward associated with this state decreases. It can be formalized with:
\begin{equation}
    R(s,a,s') = \frac{1}{\sqrt{N(s')}}
\end{equation}
where $N(s)$ is the number of times that the state $s$ has been visited. Although this method is efficient and tractable in a tabular environment (with a discrete state space), it hardly scales when states are numerous or continuous since an agent never really returns in the same state. A first solution proposed by \citet{tang2017exploration}, called \textit{TRPO-AE-hash}, is to hash the latent space of an auto-encoder fed with states. However, these results are only slightly better than those obtained with a classic exploration policy. An other line of works propose to adapt counting to high-dimensional state spaces via \textit{pseudo-counts} \cite{bellemare2016unifying}. Essentially, \textit{pseudo-counts} allow the generalization of the count from a state towards neighbourhood states using a learnt density model $\rho$. This is defined as:
%

\begin{equation}
        \hat{N}(s') = \frac{p(s'|\rho)(1-p(s'|\rho')}{p(s'|\rho')-p(s'|rho)}
\end{equation}

where $\rho'(s)$ computes the density of $s$ after having learnt on $s$. In fact,  \citet{bellemare2016unifying} show that, under some assumptions, \textit{pseudo-counts} increase linearly with the true counts. In this category, \textit{DDQN-PC} \cite{bellemare2016unifying} and 
\textit{DQN-PixelCNN} \cite{ostrovski2017count} compute $\phi$ using respectively a Context-Tree Switching model (CTS) \cite{bellemare2014skip} and a Pixel-CNN density model \cite{van2016conditional}. Although the algorithms based on density models work on environments with sparse rewards, they add an important complexity layer \cite{ostrovski2017count}. One can preserve the quality of observed exploration while decreasing the computational complexity of the pseudo-count by computing it in a learnt latent space \cite{martin2017count}. 

There exists several other well-performing tractable exploration methods like \textit{RND} \cite{burda2018exploration}, \textit{DQN+SR} \cite{machado2018count}, \textit{RIDE} \cite{ride2020roberta} or \textit{BeBold} \cite{zhang2020bebold}. These papers argue the reward they propose more or less relate to a visitation count estimation.


\paragraph{Conclusion.} Maximizing the information gain over a density model may maximize the pseudo-count, which relates to count-based objectives. They provide interesting feedbacks for exploration, but in practice, pseudo-counts are hard to approximate since they rely on a powerfull density model, a strict online estimation of density and they assume $p(s|\phi)$ strictly increases $\forall s \in S$ \cite{ostrovski2017count}. In addition, they also struggle with the problem of randomness. For instance, let us assume that one (state, action) tuple can lead to two very different states with 50\% chance each. The algorithm will manage to count for both states the number of visits, although it would take twice as long to avoid to be too much attracted. However, these methods do not address the white-noise problem since next states may be randomly generated at every steps. In this case, it is unclear how these methods could resist the temptation of going into this area since the counting associated to this state will never increase.


\subsection{Conclusion} 

\begin{table*}
\centering
\begin{threeparttable}
\begin{tabular}{|l|l|l|l|l|}
  \hline
   Method  & Stoch & Computational cost &  \multicolumn{2}{|c|}{Montezuma's Revenge}  \\
     &  &  & Score & Steps \\
  \hline
  \hline
    \textbf{Information gain over the true forward model}  & & & &  \\ 
    \hline
    PE with pixels \cite{burda2019largescale} & No & HD FM & $\sim 160$ & 200M \\
    \hline
    Dynamic-AE \cite{stadie2015incentivizing} & No & FM, AE & $0$ & 5M \\
    \hline
    PE with random features \cite{burda2019largescale} & No & FM, RE & $\sim 250$ & 100M \\
    \hline
    PE with VAE features \cite{burda2019largescale} & No & FM, VAE & $\sim 450$ & 100M \\
    \hline
    PE with ICM features \cite{burda2019largescale} & $\sim$Yes & FM, Inverse model & $\sim 160$ & 100M \\
    \cite{pmlr-v97-kim19a} &  & & $161$   & 50M \\
    \hline
    EMI \cite{pmlr-v97-kim19a} & $\sim$Yes & Large architecture & $387$ & 50M \\ 
    &  & Error model & &  \\ 
    \hline
    PE with LWM \cite{ermolov2020latent} & n/a & Whitened CL, FM & $2276$ & 50M \\
    \hline
    Learning progress \cite{schmidhuber1991curious,azar2019world}& Yes & Two forward errors & n/a & n/a \\
    \cite{lopes2012exploration,oudeyer2007intrinsic,kim2020active}  & & & & \\
    \hline
    \hline
    \textbf{Information gain over density model} & & & & \\
    \hline
    \hline
    TRPO-AE-hash \cite{tang2017exploration} & No & SimHash, AE & $75$ & 50M  \\ 
    \hline
    DDQN-PC \cite{bellemare2016unifying} & No & CTS & $3459$ & 100M\footnotemark[1]  \\ 
    \hline
    DQN-PixelCNN \cite{ostrovski2017count}& No & PixelCNN & $1671$ & 100M\footnotemark[1]   \\ 
    \hline
    $\phi$-EB \cite{martin2017count}& No & density model & $2745$ & 100M\footnotemark[1] \\
    \hline
    DQN+SR \cite{machado2018count}& n/a & Successor features & $1778$ & 20M \\ 
    & & HD FM & & \\   
    \hline
    RND \cite{burda2018exploration} & No & RE  & $8152$ & 490M\\ 
    \cite{machado2018count}& & & $524$ & 100M \footnotemark[1]  \\
    \cite{pmlr-v97-kim19a} & & & $377$ & 50M \\
    \hline
    RIDE \cite{ride2020roberta} & Yes &  FM, Inverse model & n/a & n/a \\
    & & Pseudo-count & & \\
    \hline
    BeBold \cite{zhang2020bebold} & No & Hash table & $\sim 10000$ & 2000M \footnotemark[1]\footnotemark[2]\\
    \hline
    \hline
    \textbf{Information gain over forward model} & & & & \\
    \hline
    \hline
    VIME \cite{houthooft2016vime} & Yes & Bayesian FM & n/a & n/a \\
    \hline
    AKL \cite{achiam2017surprise} & Yes & Stochastic FM & n/a & n/a \\
    \hline
    Ensemble with random features \cite{pathak2019self}& Yes & FMs, RE & n/a & n/a \\
    \hline
    Ensemble with observations \cite{yao2021sample} & Yes & FMs & n/a & n/a \\
    \hline
    Emsemble with PlaNet \cite{sekar2020planning} & Yes & FMs, PlaNet & n/a & n/a \\
    \hline
    JDRX \cite{shyam2018model} & Yes & 3 Stochastic FMs & n/a & n/a \\
    \hline
\end{tabular}
  \begin{tablenotes}
    RE: Random encoder
    PE: Prediction error.
    HD: High-dimensional.
    NN: Neural network.
    Stoch: Stochasticity.
    FM: Forward model.
    CL: Contrastive loss.
    PlaNet: \cite{hafner2018learning}
    RE: Random encoder.
    \\
    \item[1] They only provide the number of frames in the paper, we assume they do not use frame skip.
    \item[2] Result with one single seed.

  \end{tablenotes}
\end{threeparttable}

 \caption{\rebut{Comparison between different ways to maximize surprise. Stochasticity (Stoch) indicates whether the model handles the white-noise problem; $\sim$Yes means that the method may not handle the stochasticity generated by an action. Computational cost refers to highly expensive models added to standard RL algorithm. We also report the mean score on \textit{Montezuma's revenge} (Score) and the number of timesteps executed to achieve this score (Steps). We gathered results from the original paper and in other papers than the original one. Our table does not pretend to be an exhaustive comparison of methods but tries to give an intuition on their relative advantages.}}\label{tab:surprise}
\end{table*}

We detailed three ways to define and maximize the surprise of an agent, based on the expected information gain over a true model of the environment. 

\begin{figure}
    \centering
    \includegraphics[width=0.7\linewidth]{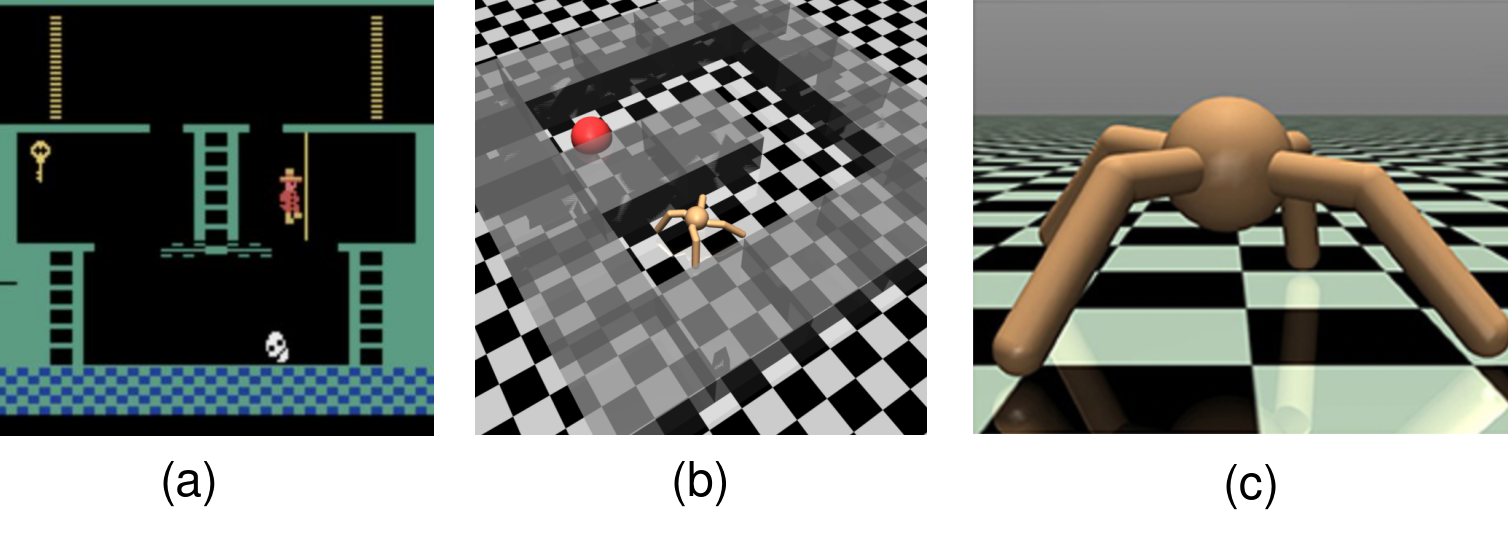}
    \caption{\rebut{Different environments widely used in our reviewed papers. (a) \textit{Montezuma's revenge}, used to assess the ability of a policy to explore. (b) \textit{Ant maze} (1x scale), used to evaluate the hierarchical organization of learnt skills (low-level: manipulation of low-level torques; high-level: navigation in the maze.) (c) \textit{Ant}, used to analyse the diversity of learnt skills.}}
    \label{fig:environments}
\end{figure}

\rebut{\tabref{tab:surprise} sums up all the surprise-based methods reviewed in this section, where it is also specified whether each method handles stochastic environments  (Stoch) (cf. \secref{sec:predictionerror}), and if expensive models are used (Computational Cost). The relative experimental advantage of each method is also reported in the \textit{Montezuma's revenge} environment (cf. Figure \ref{fig:environments}a)), a sparse-reward benchmark widely used to assess the ability of the method to explore. This gives a clue on how each method compare to the others. Methods categorized in information gain over forward model elegantly handle stochasticity from the environment, but usually apply in environments much simpler than Montezuma's revenge. We can make a similar observation with methods based on learning progress. Methods based on prediction error achieve an overall low score on Montezuma's Revenge, with stochasticity handling depending on the learnt latent space. PE with LWM \cite{ermolov2020latent} achieves good performance, presumably because the learnt representation is more appropriate. One should be cautious about the low results of the Dynamic-AE \cite{stadie2015incentivizing}, because of the very low number of timesteps. Methods based on information gain over a density model are sensitive to stochasticity since nothing prevents them to return in noisy states, but achieve overall good results on Montezuma's Revenge, thanks to the pseudo-count estimation. Among the best methods: BeBold \cite{zhang2020bebold} outstanding result has to be taken with caution, because it is not averaged over several seed; RND \cite{burda2018exploration} is a simple method that achieves important asymptotic performance.}

In practice, the expected information gain over a forward model and the learning progress well-approximate the expected information gain over the true model. Therefore, it appears that they intuitively and experimentally allow to well-explore inherently stochastic environments, but are hard to implement. The expected information gain over a density model can be seen as approximating the expected information gain over the true uniform density model. \rebut{This makes the agent targets a uniform distribution of states: while it makes the agent sensitive to stochasticity, it executes robust exploration in deterministic environment.} In fact, we discuss in the next section the relevance of aiming for a uniform distribution of states, through the study of novelty-based intrinsic motivations.

\section{Novelty maximization}\label{sec:novelty}
Novelty quantifies how much a stimuli contrasts with a previous set of experiences \cite{barto2013novelty,berlyne1966curiosity}. More formally, \citet{barto2013novelty} defend that \textit{an observation is novel when a representation of it is not found in memory, or, more realistically, when it is not “close enough” to any representation found in memory}. Previous experiences may be collected in a bounded memory or distilled in a learnt representation. 

Several works propose to formalize novelty seeking as looking for low-density states \cite{becker2021exploration}, or similarly (cf. \secref{sec:knearest}), states that are different from others \cite{lehman2011novelty,conti2018improving}. In our case, this would result in maximizing the entropy of a state distribution. This distribution can be the t-steps state distribution (cf. \eqref{eq:dpi}) $H(d^{\pi}_t(S))$ or the entropy of the stationary state-visitation distribution over a horizon $T$:
\begin{align}
 H(d^{\pi}_{0:T}(S))=H(\frac{1}{T} \sum_{t=1}^T d^{\pi}_t(S)).
\end{align} 

In practice, these distributions can be approximated with a buffer. This formalization is not perfect and does not fit several intuitions about novelty \cite{barto2013novelty}. \citet{barto2013novelty} criticize such definition by stressing out that very distinct and memorable events may have low probabilities of occurring while not being novel (\textit{e.g} a wedding). They suggest that novelty may rather relates to the acquisition of a representation of the incoming sensory data. Following this definition, we propose to formalize novelty seeking behaviors as those that \textit{actively} maximize the mutual information between states and their representation $I(S;Z)=H(S) - H(S|Z)$ where $Z$ is a low-dimensional space ($|Z| \leq |S|$). This objective is commonly known as the \textit{infomax} principle. \cite{linsker1988self,almeida2003misep,bell1995information,HjelmFLGBTB19}; in our case, it amounts to \textbf{actively} learning a representation of the environment. Most of works focus on actively maximizing the entropy of state distribution while a representation learning function minimizes $H(S|Z)$. Furthermore, if one assumes that $Z=S$, the infomax principle collapses to an entropy maximization $H(S)$. 



There are several ways to maximize the state-entropy, we separate them based on how they maximize the entropy. We found two kind of methods: low-density search and k-nearest neighbors methods.

\subsection{Direct entropy maximization}\label{sec:directdensity}

\rebut{The most evident way to maximize the entropy of states consists in maximizing $H(\rho(s))$ where $\rho(s)=p(s|\rho)$ approximates the stationary state-visitation distribution $d^{\pi}_{0:T}(S)$.} If we access this density model, it becomes straightforward to discover a policy that maximizes the entropy of a stationary state distribution \cite{hazan2019provably}. But computing $\rho(s)$ is challenging in high-dimensional state spaces. Several methods propose to estimate $\rho(s)$ using variational inference \cite{exploration2021zhang,islam2019entropy,lee2019efficient,pong2019skew} based on autoencoder architectures. 
 In this setting, we can use the VAE loss, approximated either as \eqref{eq:badapprox} \cite{vezzani2019learning,lee2019efficient} or \eqref{eq:unbiasedapprox} \cite{pong2019skew}, assuming $z$ is a compressed latent variable, $p(z)$ a prior distribution \cite{KingmaW13} and $q_{decoder}$ a neural network that ends with a diagonal Gaussian.
 
 \rebut{
 \begin{subequations}
\begin{align}
	\log \rho(s') & \geq \E_{\hat{s'} \sim q_{decoder}(\cdot|z)} - \log q_{decoder}(\hat{s'}|z) + D_{KL}(q_{encoder}(z|s)||p(z)) \\
	    &\approx - \log q_{decoder}(s'|z) + D_{KL}(q_{encoder}(z|s')||p(z)) \label{eq:badapprox}\\
		&\approx \log \frac{1}{N} \sum_{i=1}^N \frac{p(z)}{q_{encoder}(z|s')}q_{decoder}(s'|z) \label{eq:unbiasedapprox}
\end{align}
 \end{subequations}
 }
\eqref{eq:unbiasedapprox} is more expensive to compute than \eqref{eq:badapprox} since it requires decoding several samples, but presumably exhibit less variance. Basically, this estimation allows to reward an agent \cite{berseth2020smirl,lee2019efficient,exploration2021zhang} according to:
\begin{equation*}
    R(s,a,s') = - \log \rho(s').
    \label{eq:logpbs}
\end{equation*}

\citet{lee2019efficient} maximize \eqref{eq:unbiasedapprox} by learning new skills that target these novel states (see also \secref{sec:skilllearning}). \rebut{Using \eqref{eq:badapprox}, \cite{vezzani2019learning} approximates \eqref{eq:badapprox} with the ELBO as used by the VAE.} This is similar to \textit{MaxRenyi} \cite{exploration2021zhang}, which uses the Rény entropy, a more general version of the Shannon entropy, to give more importance to very low-density states. \citet{islam2019entropy} propose to condition the state density estimation with policy parameters in order to directly back-propagate the gradient of state-entropy into policy parameters. Although \textit{MaxRenyi} achieves good scores on \textit{Montezuma's revenge} with pure exploration, maximizing the ground state entropy may not be adequate since two closed ground states are not necessarily neighbors in the true environment \cite{aubret2021distop}. Following this observation, \textit{GEM} \cite{guo2021geometric} rather maximizes the entropy of the estimated density of states considering the dynamic-aware proximity of states, $H(Z)$. However they do not actively consider $H(Z|S)$.

\paragraph{Conclusion.} Generally speaking, these methods need an accurate density model to provide rewards. In the next paragraph, we study methods that avoid learning a density model.

\subsection{K-nearest neighbors approximation of entropy}\label{sec:knearest}

Several works propose to approximate the entropy of a distribution using samples and their k-nearest neighbors \cite{singh2003nearest,kraskov2004estimating}. In fact such objective has already been refered to as novelty \cite{conti2018improving}. Assuming $nn_k(S_b,s_i)$ is a function that outputs the k-th closest state to $s_i$ in $S_b$, this approximation can be written as:
\begin{equation}
	H(S) \propto \frac{1}{|S_b|} \sum_{s_i \in S_b} \log ||s_i - nn_k(S_b,s_i)||_2 + \chi(|S_b|) + Const
\label{eq:knearestequation}
\end{equation}
\begin{wrapfigure}{r}{0.5\linewidth}
\centering
\includegraphics[width=1\linewidth]{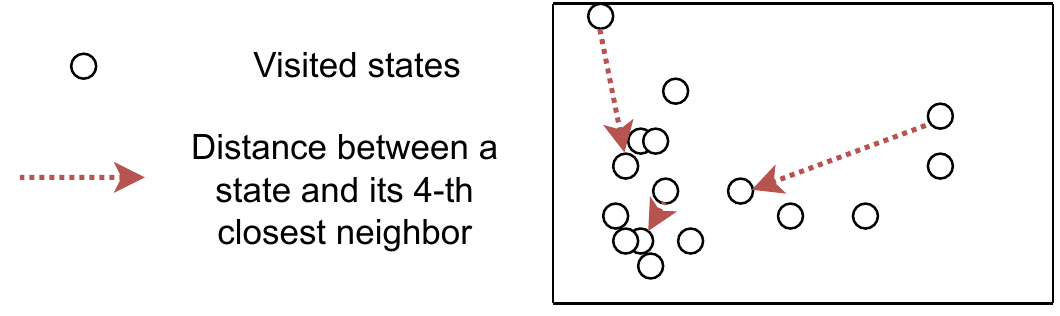}
\caption{Illustration of the correlation between density and the fourth-nearest neighbor distance.}
\label{fig:knearest}
\end{wrapfigure}

where $\chi(S_b)$ is the digamma function. This approximation assumes the uniformity of states in the ball centered on a sampled state with radius $||s_i - nn_k(S_b,s_i)||_2$ \cite{lombardi2016nonparametric} but its full form is unbiased with a large number of samples \cite{singh2003nearest}. Intuitively, it means that the entropy is proportional to the average distance between states and their neighbors. \figref{fig:knearest} shows how density estimation relates to k-nearest neighbors distance. We clearly see that low-density states tend to be more distant from their nearest neighbors. Few methods \cite{mutti2020policy} provably relates to such estimations, but several approaches take advantage of the distance between state and neighbors to generate intrinsic rewards, making them related to such entropy maximization. For instance, \textit{APT} \cite{liu2021behavior} proposes new intrinsic rewards based on the k-nearest neighbors estimation of entropy:

\begin{align}
R(s,a_t,s') = \log (1+ \frac{1}{K} \sum_0^K || f(s') - nn_k(f(S_b),f(s')) ||_2)
\end{align}

where $f$ is a representation function learnt with a contrastive loss based on data augmentation \cite{srinivas2020curl} and $K$ denotes the number of k-nn estimations. By looking for distant state embeddings during an unsupervised pre-training phase, they manage to considerably speed up task-learning in the DeepMind Control Suite. The representation $g$ can also derive from a random encoder \cite{seo2021state} or a constrastive loss that ensures the euclidean proximity between consecutive states \cite{tao2020novelty,yarats2021reinforcement}. \rebut{Alternatively, GoCu \cite{bougie2020skill} achieve SOTA results on Montezuma's revenge by learning a representation with a VAE and reward the agent based on how distant, in term of timesteps, a state is from a set of K other states.}

\paragraph{Identifying different states.}
Instead of relying on euclidean distance, one can try to learn a similarity function. \textbf{EX$^2$} \cite{fu2017ex2} learns a discriminator to differentiate states from each other: when the discriminator does not manage to differentiate the current state from those in the buffer, it means that the agent has not visited this state enough and it will be rewarded. States are sampled from a buffer, implying the necessity to have a large buffer. To avoid this, some methods distill recent states in a prior distribution of latent variables \cite{kim2019curiosity,klissarovvariational}. The intrinsic reward for a state is then the KL-divergence between a fixed diagonal Gaussian prior and the posterior of the distribution of latent variables. In this case, common latent states fit the prior while novel latents diverge from the prior.
 
\paragraph{Intra-episode novelty.} 
K-nearest neighbors intrinsic rewards have also been employed to improve intra-episode novelty \cite{stanton2018deep}. It contrasts with standard exploration since the agent looks for novel states in the current episode: typically it can try to reach all states after every resets. This setting is possible when the policy depends on all its previous interactions, which is often the case when an agent evolves in a POMDP, since the agent has to be able to predict its value function even though varies widely during episodes. This way, ECO \cite{savinov2018episodic} and Never give up \cite{badia2019never} uses an episodic memory and learn to reach states that have not been visited during the current episode. 

\paragraph{Conclusion} K-nn methods turn out to be simple to experiment, but they strongly rely on learnt dynamic-aware representations since they fully take advantage of a meaningful euclidean embedded proximity; their theoretical connection to the rigorous approximation of entropy remains most of the time unclear and the approach badly scales with an increase of the memory size. We note that simple methods can tackle the issue of finding the neighbors by partitioning together close states \cite{yarats2021reinforcement}. Overall, we observe efficient exploration and the methods easily translate to intra-episode exploration.


\subsection{Conclusion}

\begin{table*}
\centering
\begin{threeparttable}
\begin{tabular}{|l|l|l|l|l|}
  \hline
    Method  & Stoch & Computational cost &  \multicolumn{2}{|c|}{Montezuma's Revenge}  \\
     &  &  & Score & Steps \\  \hline
  \hline
    \textbf{Direct entropy maximization} & & & & \\
    \hline
    \hline
    MOBE \cite{vezzani2019learning}, StateEnt \cite{islam2019entropy} & No & VAE & n/a & n/a \\ 
    \hline
    Renyi entropy \cite{exploration2021zhang} & No & VAE, planning & $8100$ & 200M \\
     \hline
     GEM \cite{guo2021geometric} & Yes & CL & n/a & n/a \\
      \hline
    SMM \cite{lee2019efficient} & No & VAE, Discriminator & n/a & n/a \\
     \hline
    \hline      
    \textbf{K-nearest neighbors approximation of entropy} & & & & \\
    \hline 
    \hline
    EX$^2$ \cite{fu2017ex2} & No & Discriminator & n/a & n/a \\
    \cite{pmlr-v97-kim19a} & & & 0 & 50M \\
    \hline 
    CB \cite{kim2019curiosity}& Yes & IB & $\sim 1700$ & n/a \\
    \hline
    VSIMR \cite{klissarovvariational}& No & VAE & n/a & n/a \\
    \hline
    ECO \cite{savinov2018episodic} & Yes & Siamese architecture & n/a & n/a \\
    \cite{bougie2020skill} & & & $8032$ & 100M  \\
    \hline
    APT \cite{liu2021behavior} & n/a & CL & $0.2$ & 250M \\
    \hline
    RE3 \cite{seo2021state} & Yes & ensemble of REs & $100$ & 5M \\
    \hline
     Proto-RL \cite{tao2020novelty}, Novelty \cite{yarats2021reinforcement} & n/a & CL & n/a & n/a \\ 
     \hline 
     NGU \cite{badia2019never} & Yes & Inverse model, RND & $16800$ & 3500M \footnotemark[1] \\ 
     & & several policies & & \\
     \hline 
     DeepCS \cite{stanton2018deep} & No & RAM-based grid & $3500$ & 160M \\
     \hline
     GoCu \cite{bougie2020skill} & n/a & VAE, predictor & $10958$ & 100M \\
     \hline
 \end{tabular}
  \begin{tablenotes}
    NN: Neural network.
    Stoch: Stochasticity.
    FM: Forward model.
    RND: \cite{burda2018exploration}.
    IB: \cite{alemi2016deep}.
    RE: Random encoder.
    CL: Contrastive learning.
    \item[1] They only provide the number of frames in the paper, we assume they do not use frame skip.

  \end{tablenotes}
\end{threeparttable}
 \caption{\rebut{Comparison between different ways to maximize novelty. Stochasticity (Stoch) indicates whether the model handles the white-noise problem. Computational cost refers to highly expensive models added to standard RL algorithm. We also report the mean score on \textit{Montezuma's revenge} (Score) and the number of timesteps executed to achieve this score (Steps). We gathered results from the original paper and in other papers than the original one. Our table does not pretend to be an exhaustive comparison of methods but tries to give an intuition on their relative advantages.}}\label{tab:novelty}
\end{table*}
\begin{figure}
    \centering
    \includegraphics[width=1\linewidth]{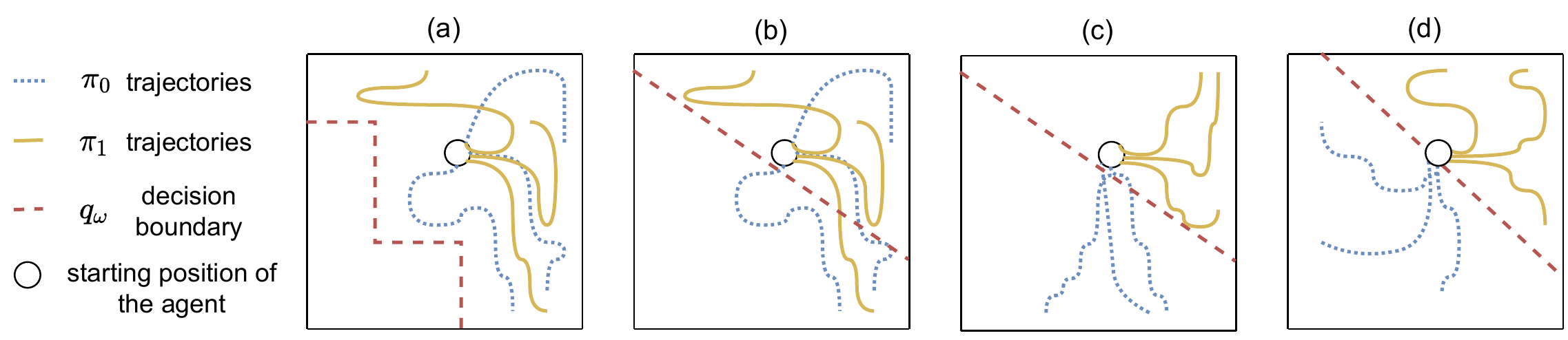}
    \caption{\rebut{Illustration of the implicit learning steps of algorithms that use a fixed goal distribution.} (a) Skills are not learnt yet. \rebut{The discriminator randomly assigns partitions of the state space to goals.} (b) The discriminator tries unsuccessfully to distinguish the skills. (c) Each skill learns to go in the area assigned to it by the discriminator. (d) Skills locally spread out by maximizing action entropy \protect\cite{haarnoja2018soft}. \rebut{The discriminator successfully partitions the areas visited by each skill.}}
    \label{fig:diaynall}
\end{figure}
In this section, we reviewed works that maximize novelty to improve exploration with flat policies. We formalized novelty as actively discovering a representation according to the infomax principle, even though most of works only maximize the entropy of states/representations of states. 

\rebut{In \tabref{tab:novelty}, we give a summary of all the novelty-based methods reviewed in this section. These methods are also compared according to their performance on the sparse reward environment \textit{Montezuma's revenge} (cf. \figref{fig:environments}a)), and whether they handle stochastic environments (cf. \secref{sec:predictionerror}). We can see that 
 these methods can better explore than surprise-based method, in particular when using intra-episode novelty mechanisms \cite{badia2019never, savinov2018episodic}. They can also be robust to stochasticity thanks to a specific learnt representation or the use of an ensemble of encoders \cite{seo2021state}.}


Works manage to learn a representation that match the inherent structure of the environment \cite{tao2020novelty}. It suggests that it is most of the time enough to learn a good representation. For instance, \citet{guo2021geometric} and \citet{tao2020novelty} compute a reward based on a learnt representation, but perhaps a bad representation tends to be located in low-density areas. It would result that active representation entropy maximization correlates with state-conditional entropy minimization. 
We are not aware of a lot of methods that actively and explicitly maximize $I(Z;S)$ in a RL. Yet, we stress out three methods that strive to actively learn a representation of states. In \textit{CRL} \cite{du2021curious} and \textit{CuRe} \cite{aljalbout2021seeking}, the agent plays a minimax game. A module learns a representation function with a constrastive loss and the agent actively challenges the representation by looking for states with a large loss.

\section{Skill learning}\label{sec:skilllearning}

In our everyday life, nobody has to think about having to move his arms' muscles to grasp an object. A command to take the object is just issued. This can be done because an acquired skill can be effortlessly reused.

Skill abstraction denotes the ability of an agent to learn a representation of diverse skills. We formalize skill abstraction as maximizing the mutual information between the goal $g \in G$ and some of the rest of the contextual states $f(\tau) \in u(\mathcal{T})$, denoted as $I(G; u(\mathcal{T}))$ where $\tau \in \mathcal{T}$ is a trajectory and $f$ a function that extracts a subpart of the trajectory (last state for example). The definition of $u$ depends on the wanted semantic meaning of a skill. Let $s_0$ refers to the state at which the skill started and $s$ a random state from the trajectory, we highlight two settings based on the literature:

\begin{itemize}
\item $u(\mathcal{T}) = S$, the agent learns skills that target a particular state of the environment \cite{eysenbach2018diversity}.
\item $u(\mathcal{T}) = \mathcal{T}$, the agent learns skills that follow a particular trajectory. This way, two different skills can end in the same state if they cross different areas \cite{co2018self}.
\end{itemize}

Most of works maximize $I(G; S)$ so that, unless stated otherwise, we refer to this objective. In the following, we will study the different ways to maximize $I(G;S)$ which can be written under its reversed form $I(S;G) = H(G) - H(G|S)$ or forward form $ I(G;S) = H(S) - H(S|G)$ \cite{campos2020explore}. In particular, we emphasize that:

\begin{subequations}
\begin{align}
- H(G | S) &= \sum_{g \in G, s \in S} p(g,s) \log p(g|s) \\
	&= \E_{\substack{g \sim p(g) \\ s \sim \pi^g  }} \log p(g|s)
	\label{eq:im}
\end{align}
\end{subequations}
where, to simplify, $p(g)$ is the current distribution of goals (approximated with a buffer) and $s \sim \pi^g$ denotes the distribution of states that results from the policy that achieves $g$. Note that $p(g,s) = p(s|g)p(g) $.

In this section, we first focus on methods that assume they can learn all skills induced by a given goal space/goal distribution and they assign parts of trajectories to every goal. The second set of methods directly derives the goal space from visited states, so that there are two different challenges that we treat separately: the agent has to learn to reach a selected goal and it must maximize the diversity of goals it learns to reach. We make this choice of decomposition because some contributions focus on only one part of the objective function.


\subsection{Fixing the goal distribution}\label{sec:predefinedG}

The first approach assumes the goal space is arbitrarily provided except for the semantic meaning of a goal. In this setting, the agent samples goals uniformly from $G$, ensuring that $H(G)$ is maximal, and it progressively assigns all possible goals to a part of the state space. To do this assignment, the agent maximizes the reward provided by \eqref{eq:im}:

\begin{equation}
R(g,s,a,s') = - \log q_{\omega}(g|s')
\label{eq:vlbim}
\end{equation}

where $q_{\omega}(g|s')$ represents a learnt discriminator (often a neural network) that approximates $p(g|s')$.

At first, we focus on discrete number of skills, where $p(g)$ represents a uniform categorical distribution. \figref{fig:diaynall} sums up the learning process with two discrete skills: 1- skills and discriminator $q_{\omega}$ are randomly initialized; 2- the discriminator tries to differentiate the skills with states $s$ from its trajectories, in order to approximate $p(g|s)$; 3- skills are rewarded with \eqref{eq:vlbim} in order to make them go in the area assigned to it by the discriminator; 4- finally, skills are clearly distinguishable and target different parts of the state space. \textit{SNN4HRL} \cite{florensa2017stochastic} and \textit{DIAYN} \cite{eysenbach2018diversity} implement this procedure by approximating $g$ with, respectively, a partition-based normalized count and a neural network. \textit{VALOR} \cite{achiam2018variational} also uses a neural network, but discriminate discrete trajectories. In this setting, the agent executes one skill per episode. 

\textit{HIDIO} \cite{zhang2020hierarchical} sequentially executes skills, yet that is not clear how they manage to avoid forgetting previously learnt skills. Maximizing $I(G;S|S_0)$ like \textit{VIC} \cite{gregor2016variational} or $I(G;S_0|S)$  with \textit{R-VIC} \cite{baumli2021relative} makes it hard to use a uniform (for instance) $H(G|S_0)$, because every skill may not be executable everywhere in the state space. Therefore, they also maximize the entropy term with another reward bonus similar to $\log p(g|s_0)$. They learn discriminable skills, but still struggle to combine them on complex benchmarks \cite{baumli2021relative}. Keeping $p(g)$ uniform, \textit{DADS} \cite{sharma2019dynamics} maximizes the forward form of mutual information $I(S;G|S_0) = H(S|S_0) - H(S|G,S_0)$ by approximating $p(s | s_0)$ and $p(s | s_0,g)$. This method makes possible to plan over skills and can combine several locomotion skills. However this requires several conditional probability density estimation on the ground state space, which may badly scale on higher-dimensional environments.

These methods tend to stay close from their starting point \cite{campos2020explore} and do not learn skills that cover the whole state space. In fact, it is easier for the discriminator to overfit over a small area than to make a policy go in a novel area, this results with a lot of policies that target a restricted part of the state space  \cite{choi2021variational}.  Accessing the whole set of true possible states and deriving the set of goals by encoding states can considerably improve the coverage of skills \cite{campos2020explore}. 

\paragraph{Approaches for a better coverage of states.} Heterogeneous methods address the problem of overfitting of the discriminator. The naive way can be to regularize the learning process of the discriminator. \textit{ELSIM} \cite{aubret2020elsim} takes advantages of L2 regularization and progressively expand the goal space $G$ to cover larger areas of the state space and  \citet{choi2021variational} propose to use spectral normalization \cite{miyato2018spectral}. More consistent dynamic-aware methods may further improve regularization; however it remains hard to scale the methods to a large number of skills which are necessary to scale to a large environment. In above-mentioned methods, the number of skills greatly increases \cite{achiam2018variational,aubret2020elsim} and the discrete skill embedding does not provide information about proximity of skills. Therefore learning a continuous embedding may be more efficient.

\paragraph{Continuous embedding.} The prior uniform distribution $p(g)$ is far more difficult to set in a continuous embedding. One can introduce the \textit{continuous DIAYN} \cite{choi2021variational,zhang2020hierarchical} with a prior $p(G) = \mathcal{N}(0^d,I)$ where $d$ is the number of dimensions, or the \textit{continuous DADS} with a uniform distribution over $[-1; 1]$ \cite{sharma2019dynamics}, yet it remains unclear how the skills could adapt to complex environments, where the prior does not globally fit the inherent structure of the environment \rebut{(\textit{e.g} a disc-shaped environment)}. \textit{VISR} \cite{visf2020ansen} seems to, at least partially, overcome this issue with a long unsupervised training phase and successor features. They uniformly sample goals on the unit-sphere and computes the reward as a dot product between unit-normed goal vectors and successor features $\log q_{\omega}(g|s) = \phi_{successor}(s)^T g$. 

\paragraph{Conclusion.} This set of methods manages to learn discrete skills that can be combined, yet, despite regularization, discrete skills struggle to cover a very large state space \cite{aubret2020elsim}. Successful adaptations to scale it up to large states spaces currently rely on the relevance of successor features. In the next two sections, we study how to maximize the mutual information by assuming the goal space derives from the state space.


%
%

\subsection{Achieving a state-goal}\label{sec:goalstate}

In this section, we review how current methods maximize the goal achievement part of the objective of the agent, $-H(S_g|S)$ where $S_g$ refers to the goal-relative embedding of states. We temporally set aside $H(S_g)$ and we will come back to this in the next subsection, \secref{eq:diversestate}, mainly because the two issues are tackled separately in the literature.

Obviously,  maximizing $- H(S_g | S)$ can be written:
\begin{align}
- H(S_g | S) &= \sum_{S_g,S} p(s_g,s) \log p(s_g|s) = \E_{\substack{s_g \sim p(s) \\ s \sim \pi^g  }} \log p(s_g|s)
\end{align}

where, to simplify, $p(s)$ is the current distribution of states (approximated with a buffer) and $s \sim \pi^g$ denotes the distribution of states that results from the policy that achieves $g$. If $\log p(s_g|s')$ is modelled as an unparameterized Gaussian with a unit-diagonal co-variance matrix, we have $\log p(s_g|s') \propto -||s_g-s'||_2^2 + Const$ so that we can reward an agent according to:
\begin{equation}
	R(s_g,s,a,s')= -||s_g-s'||_2^2.
	\label{eq:distance_reward}
\end{equation}

It means that if the goal is a state, the agent must minimize the distance between its state and the goal state. To achieve this, it can take advantage of a goal-conditioned policy $\pi^{s_g}(s)$.

\paragraph{Ground state space.} This way, \textit{Hierarchical Actor-Critic (HAC)} \cite{levy2018hierarchical} directly uses the state space as a goal space to learn three levels of option (the options from the second level are selected to fulfill the chosen option from the third level). A reward is given when the distance between states and goals (the same distance as in Equation \ref{eq:distance_reward}) is below a threshold and they take advantage of HER to avoid to directly use the threshold. Similar reward functions can be found in \citet{pitis2020maximum} and \citet{zhao2019maximum}. Related to these works, \textit{HIRO} \cite{nachum2019data} uses as a goal the difference between the initial state and the state at the end of the option $f(\Tau) = S_f - S_0$. 

This approach is relatively simple and does not require extra neural networks. However, there are two problems in using the state space in the reward function. Firstly, a distance (like L2) makes little sense in a very large space like images composed of pixels. Secondly, it is difficult to make a manager policy learn on a too large action space. Typically, an algorithm having as goals images can imply an action space of $84\times 84\times 3$ dimensions for a goal-selection policy (in the case of an image with standard shape). Such a wide space is currently intractable, so these algorithms can only work on low-dimensional state spaces. 

\paragraph{Learning a representation of goals.} To tackle this issue, an agent can learn low-dimensional embedding of space $\phi_e$ and maximize the reward of \eqref{eq:distance_reward_phi} using a goal-conditioned policy $\pi^{f(s_g)}(s)$:

\begin{equation}
	R(s_g,s,a,s')= -||f(s_g)-f(s')||_2^2.
	\label{eq:distance_reward_phi}
\end{equation}

Similarly to \eqref{eq:distance_reward}, this amounts to maximize $- H(f(S_g) | f(S))$. \textit{RIG} \cite{nair2018visual} proposes to build the feature space independently with a variational auto-encoder (VAE); but this approach can be very sensitive to distractors (i.e. useless features for the task or goal, inside states) and does not allow to correctly weight features. Similar approaches also encode part of trajectories \cite{kim2021unsupervised,co2018self} for similar mutual information objectives. \textit{SFA-GWR-HRL} \cite{zhou2019vision} uses unsupervised methods like the algorithms of \textit{slow features analysis} \cite{wiskott2002slow} and \textit{growing when required} \cite{marsland2002self} to build a topological map. A hierarchical agent then uses nodes of the map, representing positions in the world, as a goal space. However the authors do not compare their contribution to previous approaches. 
 
Other approaches learn a state embedding that captures the proximity of states with contrastive losses. For instance, \textit{DISCERN} learns the representation function by maximizing the mutual information between the last state representation and the state-goal representation. Similarly to works in \secref{sec:predefinedG}, the fluctuations around the objective allow to bring states around $s_g$ closer to it in the representation. More explicitly, the representation of \textit{NOR} \cite{nachum2019near} maximizes $I(f(S_{t+k});f(S_t),A_{t:t+k})$ and the one of \textit{LESSON} \cite{li2021learning} maximizes $I(f(S_{t+1});f(S_t))$; \textit{LESSON } and \textit{NOR} target a change in the representation and manage to navigate in a high-dimensional maze while learning the intrinsic Euclidean structure of the mazes (cf. \tabref{tab:skills}). Their skills can be reused on several environments. However, experiments are made in 2-dimensional embedding spaces and it remains unclear how relevant may be goals as state changes in an embedding space with higher dimensions. The more the number of dimensions increase, the more difficult it will be to distinguish possible skills from impossible skills in a state. \rebut{In addition, they need dense extrinsic rewards to learn to select the skills to execute. Thus, they generate tasks with binary rewards at a location uniformly distributed in the environment such that the agent learn to achieve the tasks from the simplest to the hardest. This progressive learning generates a curriculum, helping to achieve the hardest task.} 

\paragraph{Conclusion.} To sum up, representation learning methods allows to learn state-based skills over complex state spaces. Learning this representation function combined with the use of the euclidean distance as reward function amounts to learn a particular form of reward function in addition for providing pre-computed features to the goal-conditioned policy. \rebut{As highlighted by \tabref{tab:skills}, learnt representations allow to scale the approaches to more complex goal spaces}. In the next paragraph, we study how to maximize $H(S)$ so that to make sure learnt skills target different areas of the state space. \rebut{As highlighted by \tabref{tab:skills}, it will make possible to reach very distant goals without being assisted by a curriculum of tasks.}

\subsection{Proposing diverse state-goals}\label{eq:diversestate}

To make sure the agent maximizes the mutual information between its goals and all visited states, it must sample a diverse set of goal-states. In other words, it has to maximize $H(S_g)$ but through goal selection rather than with an intrinsic bonus as in \secref{sec:novelty}. Similarly to works on novelty (cf. \secref{sec:novelty}), such entropy maximization along with skill acquisition (cf. \secref{sec:goalstate}) tackles the exploration challenge, but without facing catastrophic forgetting (cf. \secref{sec:detachment}) since the agent does not forget its skills.

A naive approach would be to generate random values in the goal space, but this faces a considerable problem: the set of achievable goals is often a very small subset of the entire goal space. To tackle this, a first approach can be to explicitly learn to differentiate these two sets of goals \cite{florensa2018automatic,racaniere2019automated}, using for example a Generative Adversarial Networks (GAN) \cite{florensa2018automatic,goodfellow2014generative}, but it is ineffective in complex environments \cite{pong2019skew}. Other works obtain good results on imagining new goals, but using a compoundable goal space, given \cite{colas2019curious} or learnt with a dataset \cite{khazatsky2021can}; results show it may be a strong candidate for object-based representations. In contrast, in a more general case, an agent can simply set a previously met state as a goal, this way, it ensures that goals are reachable, since they have already been achieved. In the rest of this section, we focus on this set of methods.%

\begin{figure}
\centering
\includegraphics[width=0.8\linewidth]{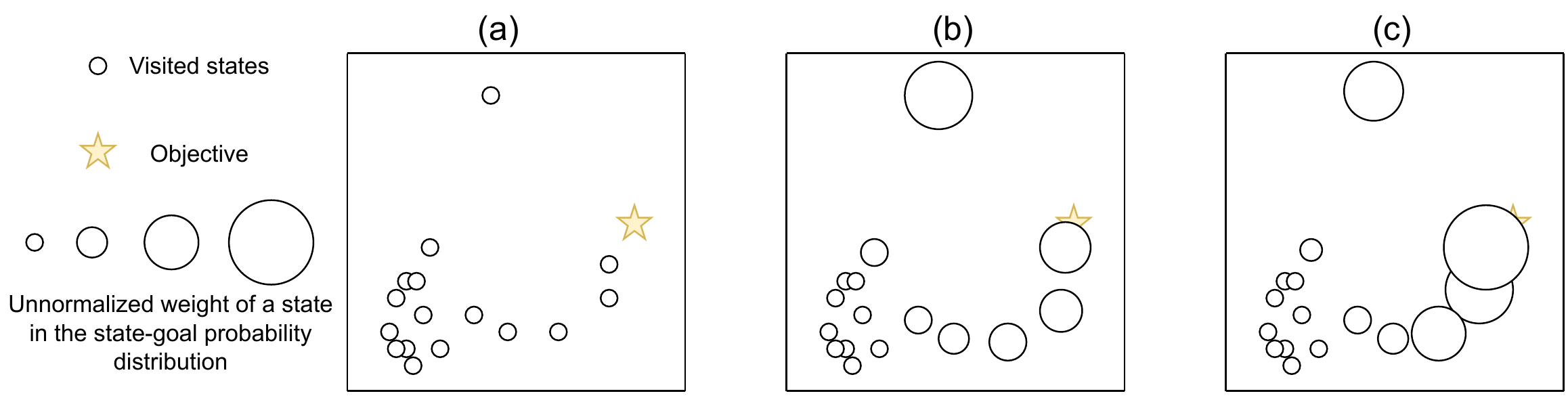}
\caption{\rebut{Illustration of the reweighting process. (a) probability of visited states to be selected as goals before reweighting; (b) probability of visited states to be selected as goals after density reweighting; (c) probability of visited states to be selected as goals after density/reward reweighting. This figure completes and simplifies the figure of \protect\citet{pong2019skew}.}}
\label{fig:reweight} 
 
\end{figure}

 In \textit{RIG} \cite{nair2018visual}, the agent randomly samples states as goals from its buffer, but it does not increase the diversity of states, and thus, the diversity of learnt skills. \citet{pong2019skew} showed theoretically and empirically that, by sampling goals following a $\alpha$-more uniform distribution over the support of visited states than the "achieved" distribution, the distribution of states of the agent can converge to the uniform distribution. Intuitively, the agent just samples more often low-density goals as illustrated it in \figref{fig:reweight}. There are several ways to increase the importance of low-density goal-states that we introduce in the following.
 
\paragraph{Density estimation in the ground state space.}  \textit{DISCERN} \cite{warde2018unsupervised} proposes to sample uniformly over the support of visited stated with a simple procedure. Every time the agent wants to add an observation to its buffer, it randomly samples an other observation from its buffer and only keeps the one that is the farthest to all other states of the buffer. This way, it progressively builds an uniform distribution of states inside its buffer. However, it uses the euclidean distance to compare images, which may not be relevant. Other approaches select the state that has the lower density (\textit{OMEGA}) \cite{pitis2020maximum} according to a kernel density estimation or use the rank of state-densities \cite{zhao2019curiosity} estimated with a Variational Gaussian Mixture Model \cite{blei2006variational}. In contrast with them, \textit{Skew-fit} \cite{pong2019skew} provides more flexibility on how uniform one want its distribution of states. \textit{Skew-fit} extends RIG and learns a parameterized generative model $q_{\rho}(S) \approx p(S)$ and skews the generative model (VAE) with the ratio:
\begin{equation}
	q_{\rho}(s)^{\alpha_{skew}} \label{eq:skewratio}
\end{equation}

 where $\alpha_{skew} < 0$ determines the speed of uniformisation. This way it gives more importance to low-density states. Then it weights all visited states according to the density approximated by the generative model at the beginning of each epoch, which is made of a predefined number of timesteps. Skew-fit manages to explore image-based environments very efficiently. As highlighted in \cite{aubret2021distop}, this ratio applied on a discrete number of skills, amount to rewards a Boltzmann goal-selection policy with:
 
 \begin{equation}
 	R(s_g) = (1+\alpha_{skew}) \log p(s_g).
 \end{equation}

\paragraph{Density reweighting by partitioning the embedding space.} With a different objective, \textit{GRIMGREP} \cite{kovavc2020grimgep} partitions the VAE embedding of Skew-fit with a Gaussian Mixture Model \cite{rasmussen1999infinite} to estimate the learning progress of each partition and avoid distractors. The density weighting can also operate in a learnt embedding. \textit{HESS} \cite{li2021efficient} partitions the embedding space of \textit{LESSON} and rewards with a variant of a count-based bonus (see \secref{sec:infogain}). It improves exploration in a two-dimensional latent embedding but the size of partitions may not scale well if the agent considers more latent dimensions. In contrast, \textit{DisTop} \cite{aubret2021distop} dynamically clusters a dynamic-aware embedding space using a variant of a Growing When Required \cite{marsland2002self}; they estimate the density of state according to how much its partition contains states and skew the distribution of sampled similarly to Skew-fit. \textit{HESS} and \textit{DisTop} demonstrate their ability to explore and navigate with an ant inside complex mazes without extrinsic rewards. \rebut{As shown in \cite{aubret2021distop} (illustration in \figref{fig:reweight}c), it is also possible to use extrinsic rewards to weight the distribution of sampled state-goals.}

\paragraph{Conclusion.} Entropy maximization methods improves over standard skill learning methods by learning to reach as many states as possible. 
We expect further works to show the ability to scale to even more complex environments, with higher-dimensional latent structure. For example, learning compositional representations ( modeling disentangled objects and relations) remains hard: \rebut{SOTA methods only manipulate few objects \cite{pong2019skew}.} 

\subsection{Conclusion} 

\begin{table*}
\centering
\begin{threeparttable}
\begin{tabular}{|l|l|l|l|l|l|}
    \hline
    \multicolumn{6}{|l|}{\textbf{Fixing the goal distribution} }\\
    \hline
    \hline
     \multicolumn{6}{|l|}{SNN4HRL \cite{florensa2017stochastic}, DIAYN \cite{eysenbach2018diversity}, VALOR \cite{achiam2018variational}}\\
    \multicolumn{6}{|l|}{HIDIO \cite{zhang2020hierarchical}, R-VIC \cite{baumli2021relative}, VIC \cite{gregor2016variational} }\\

     \multicolumn{6}{|l|}{DADS \cite{sharma2019dynamics}, continuous DIAYN \cite{choi2021variational}, ELSIM \cite{aubret2020elsim} }\\
     \multicolumn{6}{|l|}{VISR \cite{visf2020ansen}}\\
     \hline
    \hline     
  \hline
   Methods &  Scale & Goal space & Curriculum & Score & Steps\\
  \hline
  \hline
    \textbf{Achieving a state-goal}  & \multicolumn{5}{|c|}{}  \\
    \hline 
    \hline
    HAC \cite{levy2018hierarchical} & n/a & n/a & n/a & n/a & n/a \\
    \hline
    HIRO \cite{nachum2019data} & 4x & (x,y) & Yes & $\sim 0.8$ & 4M \\
    \hline
    RIG \cite{nair2018visual} & n/a& n/a & n/a  & n/a  & n/a \\
    \hline
    SeCTAR \cite{co2018self} & n/a & n/a & n/a & n/a & n/a \\
    \hline
    IBOL \cite{kim2021unsupervised} & n/a& n/a & n/a  & n/a & n/a \\
    \hline
    SFA-GWR-HRL \cite{zhou2019vision}& n/a& n/a & n/a  & n/a & n/a \\
    \hline
    NOR \cite{nachum2019near} &  4x & T-V + P& Yes & $\sim 0.7$ & 10M \\
    \cite{li2021learning}  & 4x& T-V + P & Yes & $\sim 0.4$ & 4M \\
    \hline
    LESSON \cite{li2021learning} & 4x & T-V + P & Yes & $\sim 0.6$ & 4M\\
     \hline
     \hline
     \textbf{Proposing diverse state-goals} &  \multicolumn{5}{|c|}{}  \\
    \hline
    \hline
    Goal GAN \cite{florensa2018automatic}  & 1x & (x,y) & No & Coverage $0.71\%$ & n/a \\
    \cite{pong2019skew}& n/a (\~16x) & (x,y) & No & All goals distance $\sim 7$ & 7M \\ 
    \hline
    FVC \cite{racaniere2019automated} & n/a & n/a & n/a & n/a & n/a \\
    \hline
    Skew-fit \cite{pong2019skew}& n/a (\~16x) & (x,y) & No & All goals distance $\sim 1.5$ & 7M \\ 
    \hline
    DisTop \cite{aubret2021distop} & 4x & T-V& No & $\sim 1$ & 2M \\
    \hline
    DISCERN \cite{warde2018unsupervised} & n/a & n/a & n/a & n/a & n/a \\
    \hline
    OMEGA \cite{pitis2020maximum}  & 4x & (x,y) &  No & $\sim 1$ & 4.5M \\
    \hline
     CDP \cite{zhao2019curiosity} & n/a & n/a & n/a & n/a & n/a \\
     \cite{pong2019skew}& n/a (\~16x) & (x,y) & No & All goals distance $\sim 7.5$ & 7M \\ 
     \hline
    GRIMGREP \cite{kovavc2020grimgep} & n/a& n/a & n/a  & n/a & n/a \\
    \hline
    HESS \cite{li2021efficient} & 4x & T-V + P& No & success rate $\sim 1$ & 6M \\
     \hline
 \end{tabular}
\end{threeparttable}
 \caption{\rebut{Summary of papers that learn skills through mutual information maximization. We selected the Ant Maze environment to compare methods since this is the most commonly used environment. We did not find a common test setting allowing for a fair comparison of methods in "Fixing the goal distribution''. The scale refers to the size of the used maze. Goal space refers to the \textit{a priori}  state space used to compute goals, from the less complex to the more complex: (x,y); 75-dimensional top-view of the maze (T-V); top-view + proprioceptive state (T-V + P). Curriculum refers to whether the authors use different goal locations during training, creating an implicit curriculum that makes easier learning to reach distant goals from the starting position. The score, unless stated otherwise, refers to the success rate in reaching the farthest goal.}}\label{tab:skills}
\end{table*}

We found two main ways to discover skills. The first one provides a goal space and assigns goals to areas of the state space. There are empirical evidences emphasizing that it struggles to learn and sequentially executes skills that target different areas of the state space. The second method derives the goal space from the state space with a representation learning method and over-weights the sampling of low-density visited areas. \rebut{This set of works showed the ability to hierarchically navigate in simple environments using moderately morphologically complex agents. }

\rebut{In \tabref{tab:skills}, we synthesize the methods presented in this section. We also compare skill learning methods according to their performance on the widely used hierarchical task \textit{Ant maze} (cf. \figref{fig:environments}b)), and whether they need a hand-made goal space (x,y) or an implicit curriculum of objectives. We can make two major observations: 1- methods that do not propose diverse goal-states require an implicit curriculum to learn the Ant-Maze task \cite{nachum2019data,li2021learning} (\textit{Curriculum} column); 2- contrastive representations seem crucial to avoid using a hand-defined goal space like the (x,y) coordinated (\textit{Goal space} column) \cite{nachum2019near,li2021efficient}. For methods in the "fixing the goal distribution'', we did not find a representative and widely used evaluation protocol/environment among works. However, as an example, several qualitative analysis emphasize the diversity of behaviors that can be learnt by the ant displayed in \figref{fig:environments}c) \cite{sharma2019dynamics,eysenbach2018diversity}.
}

\section{Outlooks of the domain}\label{sec:outlooks}

In this section, we take a step back and thoroughly analyze the results of our overall review. We first study the exploration process of flat intrinsic motivation in comparison with hierarchical intrinsic motivations in \secref{sec:detachment}; then, this will motivate our focus on the challenges induced by learning a deep hierarchy of skills in \secref{sec:dev}. Finally, in \secref{sec:flatim}, we discuss how flat and hierarchical intrinsic motivations can and should cohabit in such hierarchy.

\subsection{Long-term exploration, detachment and derailment}\label{sec:detachment}

\begin{figure}
\centering
\includegraphics[width=0.7\linewidth]{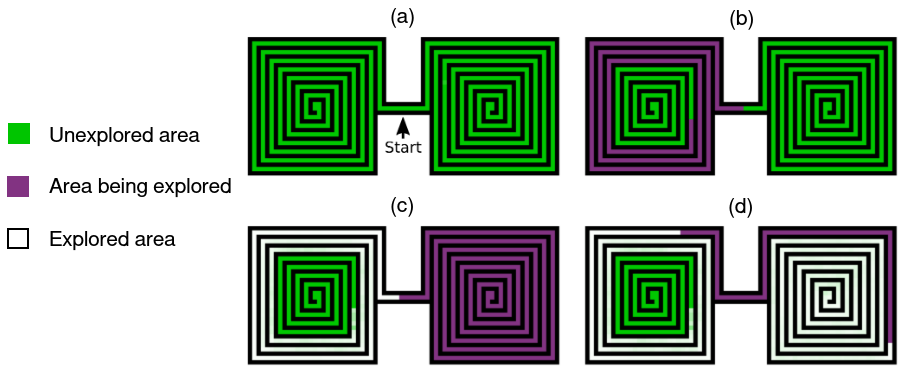}
\caption{Illustration of the \textit{detachment} issue. Image extracted from \protect\citet{goexplore}. Green color represents intrinsically rewarding areas, white color represents no-rewards areas and purples areas are currently being explored. (a) The agent has not explored the environment yet. (b) It discovers the rewarding area at the left of its starting position and explores it. (c) It consumed close intrinsic rewards on the left part, thus it prefers gathering the right-part intrinsic rewards. (d) Due to catastrophic forgetting, it forgot how to reach the intrinsically rewarding area on the left.}
\label{fig:detachment2}
\end{figure}
 
The most challenging used benchmarks in flat intrinsic motivations (surprise and novelty) are \textit{DMLab} and \textit{Montezuma's revenge}, yet very sparse reward games such as \textit{Pitfall!} are not currently addressed and should be investigated. In \textit{Pitfall!}, the first reward is reached only after multiple rooms where it requires specific action sequences to go through each room. State of the art on IM methods \cite{ostrovski2017count} achieve 0 mean reward in this game. At the opposite, imitation RL methods \cite{aytar2018playing,hester2018deep} are insensitive to such a specific reward, and thus, exceed IM methods with a mean reward of 37232 on \textit{Montezuma's revenge} and 54912 on \textit{Pitfall!}. Even though these methods use expert knowledge, this performance gap exhibits their resilience to long-term rewards. Compared with flat intrinsic reward methods, which do not exceed a 10000 score on \textit{Montezuma's revenge} \cite{burda2018exploration} and hardly achieve a score on \textit{Pitfall!} \cite{ostrovski2017count}, it shows that flat IMs is still far from solving the overall problem of exploration.

Furthermore, we want to emphasize that the challenge is harder when the intrinsic reward itself is sparse \cite{burda2018exploration}. In \textit{Montezuma's revenge}, it is about avoiding to use a key too quickly in order to be able to use it later. In every day life, it can be about avoiding to spend money too quickly. In fact, it looks like there is an exploration issue in the intrinsic reward function. Intrinsic reward can guide the exploration at the condition that the agent finds this intrinsic reward. There may be two reasons causing the intrinsic reward to be sparse:

\begin{enumerate}
\item The first comes from partial observability, with which most models are incompatible. Typically, if an agent has to push a button and can only see the effect of this pushing after a long sequence of actions, density models and predictive models may not provide meaningfull intrinsic rewards. There would be a too large distance between the event "push a button" and the intrinsic reward.
\item \figref{fig:detachment2} illustrates the second issue, called \textit{detachment} \cite{goexplore,ecoffet2021first}. It results from a distant intrinsic reward coupled with catastrophic forgetting. Simply stated, the RL agent can forget the presence of an intrinsic reward in a distant area: this is hard to maintain the correct Q-value that derives from a distant currently unvisited rewarding area. This is emphasized in on-policy settings.
\end{enumerate} 

Pursuing such distant intrinsic reward may be even harder due to the possible \textit{derailment} issue \cite{goexplore,ecoffet2021first}. Essentially, an agent may struggle to execute a long sequence of specific actions needed to reach a distant rewarding area because the local stochasticity incites local dithering all along the sequence. Detachment motivates the need for a hierarchical exploration  \cite{ecoffet2021first} and derailment motivates frontier-based exploration \cite{bharadhwaj2020leaf}, which consists in deterministically reaching the area to explore before starting exploration.



\subsection{Deeper hierarchy of skills}\label{sec:dev}

 According to \citet{brooks1991intelligence}, \textit{everything is grounded in primitive sensor motor patterns of activation}. This \textit{everything} may refer to the structure of the world and agent affordances. Capturing this knowledge amounts to form concept representations and reusable skills \cite{weng2001autonomous}, use it as a basis for new skills \cite{prince2005ongoing}, explore the environment to find new interesting skills, autonomously self-generate goals in accordance with the level and morphology of the agent.
 
 
Most works presented in \secref{sec:skilllearning} abstract actions on a restricted number of hierarchies (generally one hierarchy). This is necessary to well-understand the mechanism of abstraction, but we want to argue that imposing deeper hierarchies could considerably enhance the semantic comprehension of the environment of an agent. Organisms are often assumed to deal with composition of behaviors, which in turn serve as building block for more complex behaviors \cite{flash2005motor}. This way, using a limited vocabulary of skills makes easier avoiding the curse of dimensionality associated to the redundancy of a whole set of ground behaviors. 

 Our surveyed works \cite{nachum2019near,aubret2021distop,li2021learning,guo2021geometric,ermolov2020latent} already propose to learn the representations using the slowness principle \cite{wiskott2002slow} which assumes temporally close states should be similarly represented. By configuring the time-extension of the representation, one may focus on different semantic parts of the state space. This can be seen in \secref{sec:abstraction}: 1- the agent can learn a very low level representation that provides skills that can manipulate torques of a creature \cite{aubret2021distop}; 2- skills can also orientate an agent in a maze by extracting (x,y) coordinates from a complex state representation \cite{li2021efficient}. While they do not try to combine and learn several representations at the same time, further works could consider separate different parts of states (\textit{e.g.} agent positions and object positions \cite{mutual2021zhao}) or learning these representations at different time scales. In practice, data-augmentation methods already allow to learn object-oriented representations \cite{mitrovic2020representation,grill2020bootstrap,mussa2004neural}. Most augmentations could also be derived with contrast over time by considering, for instance, an embodied agent moving its eyes/head (crops), turning its head (rotation), controlling vergence (blur) or, without interventions, color and brightness changes \cite{chen2020simple}. Overall, it stresses out the potential of time-contrastive representations for disentangling the whole state space and providing semantically different skills; new works in this area may unlock new kind of skills. 
 
 \textit{Skill focus.}
In a developmental process, multi-level hierarchical RL questions the ability of the agent to learn all policies of the hierarchy simultaneously. This obviously relates to the ability of organisms to continually learn throughout their lifetime; but in more practical way, it may allow to focus the learning process of skills that are interesting for higher-level skills. This focus avoids learning everything in the environment \cite{aubret2021distop}, which is hard and obviously not done by biological organisms. For instance, most persons can not do a somersault. 

\textit{Critical periods and lifelong learning.}
Considering a goal representation that changes over time introduces new issues for the agent. In this case, the goal-conditioned policy may be perturbed by the changes of inputs and may no longer be able to reach the goal \cite{li2021efficient}. Current methods consider 1- developmental periods (unsupervised pre-training \cite{metzen2013incremental}); 2- to modify the representation every k-steps epochs \cite{pong2019skew}; 3- to impose slowly changes of the representation \cite{li2021efficient}. Further works may thoroughly investigate the relation and transitions between these methods since they can relate to the concept of critical periods \cite{hensch2004critical,konczak2004neural}. Critical periods assume that the brain is more plastic at some periods of development in order to acquire specific knowledge. Despite this mechanism, the brain slowly keeps learning throughout the lifetime. In the hierarchy of skills, the introduction of a new level may first result in a quick/plastic learning process, followed by slower changes. 

\subsection{The role of flat intrinsic motivations}\label{sec:flatim}

In \secref{sec:detachment}, we essentially criticized the limited role that flat intrinsic motivation like surprise or novelty can play in favor of exploration and we hypothesized in \secref{sec:dev} that deeper hierarchies could make emerge an understanding of more complex affordances. Then, what could be the roles of surprise and novelty ?

\textit{Novelty.} We saw in \secref{sec:novelty} that novelty seeking behaviors allow to learn a correct representation of the whole environment; this can be a basis for learning diverse skills. While some methods consider a goal as a state and manage to avoid using novelty bonuses \cite{pong2019skew}, this is harder to do when skills have a different semantic (like a change in the state space). \citet{nachum2019near} provide a meaningful example of this: the agent acts to simultaneously discover a representation of the environment and achieve upper-level goals.

\textit{Surprise.} We leave aside the interest of surprise for learning a forward model that could be used for planning \cite{hafner2019learning} and rather focus on the learning process. Surprise amounts to look for the learning progress of forward models so that, in a hierarchy of skills, it quantifies whether skills can currently be better learnt or not. This links surprise to curriculum learning \cite{bengio2009curriculum}, \textit{i.e} can we find a natural order to efficiently learn skills ? For example, assuming an agent want to learn to reach state-goal in a maze, it would be smarter to learn to start learning skills that target goals close to its starting position and to progressively extend its goal selection while learning other skills. Several strategies have been proposed to smartly hierarchically select goals \cite{colas2019curious,linke2020adapting}, yet it often does not consider intrinsic skills \cite{colas2019curious}.

To sum up, we propose that the role of surprise and novelty may rather be to support the learning of skills. Novelty seeking helps to learn the representation required by the skill learning module and surprise speeds up the maximization of the skill learning objective. They may interact as a loop: first, the agent learns a new representation, then it evaluates surprise to select which skill to improve and the skill learning process starts. Considering this, it would result several surprises and novelties: an agent can experiment a novel or surprise interaction for a level of decision (injure the toy while walking), yet it does not mean other levels would be surprised (it is still on the same road). This emphasizes the multi-dimensionality and relativity of the notion of surprise ou novelty \cite{berlyne1960conflict}, only a part of the incoming stimuli may arouse the agent.

\section{Conclusion}

In this survey, we have presented the current challenges faced by DRL: namely 1- learning with  \textit{sparse rewards} through exploration; 2- \textit{building a hierarchy of skills} in order to make easier credit assignment, exploration with \textit{sparse rewards} and \textit{transfer learning}.

We identified several types of IM to tackle these issues, that we classified into three categories based on a maximized information theoretic objective, which are \textit{surprise}, \textit{novelty} and \textit{skill learning}. Surprise and novelty based intrinsic motivations implicitly improve flat exploration while skill learning allows to create a hierarchy of reusable skills that also improve exploration. 

\textbf{Surprise} results from maximizing the mutual information between the true model parameters and the next state, knowing the previous state, the action and the history of interactions. We have shown that it can be maximized through three set of works: information gain over predictive models, over density models or prediction errors/learning progress. In practice, we found that the information gain over density model is ill-defined for purely stochastic areas and that the determinism assumption underpinning prediction error methods complicates their application. \rebut{Good approximations of surprise are notably useful to allow exploration in stochastic environments.} Next challenges may be to make good approximations of surprise tractable. 

\textbf{Novelty} seeking can be assimilated to learning a representation of the environment, through the maximization of mutual information between states and their representation. The most important term to actively maximize looks to be the entropy of state or representation, which can be approximated in two ways: 1- one can reward according to the parametric density of its next state, but it is complicated to estimate; 2- one can also reward an agent according to the distance between a state and currently already visited states, making the approach tractable in particular when the agent learns a dynamic-aware representation. \rebut{We found these methods to achieve state-of-the-art performance on the hard exploration task Montezuma's Revenge.} We expect future works to benefit from directly looking for good representations rather than uniformity of states.

Finally, using \textbf{skill learning} objective that amount to maximize the mutual information between a goal and a part of trajectories of the corresponding skill, an agent can learn hierarchies of temporally-extended skills. Skills can be directly learnt by attributing part of a fixed goal space to areas, but it remains to clarify how well goals can be embedded in a continuous way and whether approaches may be robust when skills are sequentially executed. The second approach derives the goals space from the state space, often through a time-contrastive loss, and expand the skill set by targeting low-density areas. \rebut{These methods manage to explore an environment while being able to return to previously visited areas}. It remains to be demonstrated how one could create larger hierarchies of skills.

The three objectives are compatible and we have discussed how they could interact to provide a robust exploration with respect to the \textit{detachment} issue, along with reusable hierarchical skills, a quick and focused skill acquisition and multi-semantic representations. 
\bibliographystyle{ACM-Reference-Format}
\bibliography{references}

\newpage
\appendix

\section{Notations}\label{app:notations}

\begin{table*}[h]
\centering
\begin{tabular}{|c|c|}
    \hline
    $\propto$ & Proportional to \\
  \hline  
  $x \sim p(\cdot)$ & $x \sim p(x)$ \\
    \hline
    $|| x ||_2$ & Euclidian norm of $x$ \\ 
    \hline
    $t$ & timestep \\
    \hline 
    $Const$ & arbitrary constant \\
     \hline
      $A$ & set of possible actions \\
      \hline
      $S$ & set of possible states \\
  \hline
      $a \in A$ & action \\
  \hline  
      $s \in S$ & state \\
\hline  
  $s_0 \in S$ & first state of a trajectory \\
  \hline  
  $s_f \in S$ & final state of a trajectory \\
  \hline  
      $s' \in S$ & state following a tuple $(s,a)$ \\
  \hline  
      $h$  & history of interactions $(s_0,a_0,s_1,\dots)$\\
  \hline
    $\hat{s}$ & predicted states \\
    \hline
    $g \in G$ & goal \\
  \hline
    $s_g \in S$ & state used as a goal \\
  \hline 
    $S_b$ & set of states contained in $b$ \\
  \hline
    $\tau \in \mathcal{T}$ & trajectory \\
\hline
    $u(\tau)$ & function that extracts parts of the trajectory $\tau$ \\
  \hline  
      $R(s,a,s')$ & reward function \\
  \hline
    $d^{\pi}_t(s)$ & t-steps state distribution \\
  \hline
  $d^{\pi}_{0:T}(S)$ &  stationary state-visitation distribution of $\pi$ over a horizon T \\
   & $\frac{1}{T} \sum_{t=1}^T d^{\pi}_t(S))$\\
    \hline  
      $f$ & representation function \\
  
  \hline 
    $z$ & compressed latent variable, $z=f(s)$ \\
    \hline  
      $\rho \in \Rho$ & density model \\
  \hline 
    $\phi \in \Phi$ & forward model \\
  \hline 
   $\phi_T \in \Phi_T$ & true forward model \\
  \hline
    $q_{\omega} $ & parameterized discriminator \\
  \hline  
      $\pi$ & policy \\
  \hline  
      $\pi^g$ & policy conditioned on a goal $g$ \\
  \hline  
      $nn_k(S,s')$ & k-th closest state to $s'$ in $S$  \\
  \hline
  $D_{KL}(p(x) || p'(x))$ & Kullback–Leibler divergence \\
   & $\E_{x \sim p(\cdot)} \log \frac{p(x)}{p'(x)}$ \\
  \hline
  $H(X)$ & $-\int_{X} p(x)\log p(x)$ \\
  \hline
  $H(X|S)$ & $-\int_{S} p(s)\int_{X} p(x|s)\log p(x|s) dx ds$ \\
  \hline
  $I(X;Y)$ & $H(X) - H(X|Y)$\\
  \hline
  $I(X;Y|S)$ & $H(X|S) - H(X|Y,S)$ \\
  \hline
  $IG(h,A,S',S,\Phi)$ & Information gain \\
   & $I(S';\Phi|h,A,S)$ \\
  \hline
 \end{tabular}
 \caption{Notations used in the paper.}\label{tab:notations}
\end{table*}

\end{document}